\documentclass[journal]{IEEEtran}
\pdfoutput=1
\usepackage{amsmath,amsfonts}
\usepackage{algorithmic}
\usepackage{array}
\usepackage[caption=false,font=normalsize,labelfont=sf,textfont=sf]{subfig}
\usepackage{textcomp}
\usepackage{stfloats}
\usepackage{url}
\usepackage{verbatim}
\usepackage{graphicx}
\usepackage{makecell}
\def\BibTeX{{\rm B\kern-.05em{\sc i\kern-.025em b}\kern-.08em
    T\kern-.1667em\lower.7ex\hbox{E}\kern-.125emX}}
\usepackage{balance}

\begin{document}
\title{A Robust Document Image Watermarking Scheme using Deep Neural Network}
\author{
	Sulong~Ge,
	Zhihua~Xia,
	Jianwei~Fei,
	Xingming~Sun,
	and Jian~Weng
  \thanks{
		\IEEEcompsocthanksitem Sulong Ge, Jianwei Fei, and Xingming Sun are with Engineering Research Center of Digital Forensics, Ministry of Education, School of Computer and Software, Jiangsu Engineering Center of Network Monitoring, Jiangsu Collaborative Innovation Center on Atmospheric Environment and Equipment Technology, Nanjing University of Information Science \& Technology, Nanjing, 210044, China.
		\IEEEcompsocthanksitem Zhihua Xia and Jian Weng are with College of Cyber Security, Jinan University, Guangzhou, 510632, China. Zhihua Xia is also with Engineering Research Center of Digital Forensics, Nanjing University of Information Science \& Technology, Nanjing, 210044, China.
		\IEEEcompsocthanksitem Zhihua Xia and Jian Weng are the corresponding authors. e-mail: xia\_zhihua@163.com, cryptjweng@gmail.com. 
		\IEEEcompsocthanksitem The code of our scheme can be found at https://github.com/gslxr/Document-image-watermarking
	}
}

\markboth{Journal of \LaTeX\ Class Files,~Vol.~18, No.~9, September~2020}%
{Ge \MakeLowercase{\textit{et al.}}:Robust Document Image Watermarking using Deep Neural Network}

\maketitle

\begin{abstract}
	Watermarking is an important copyright protection technology which generally embeds the identity information into the carrier imperceptibly. Then the identity can be extracted to prove the copyright from the watermarked carrier even after suffering various attacks. Most of the existing watermarking technologies take the nature images as carriers. Different from the natural images, document images are not so rich in color and texture, and thus have less redundant information to carry watermarks. This paper proposes an end-to-end document image watermarking scheme using the deep neural network. Specifically, an encoder and a decoder are designed to embed and extract the watermark. A noise layer is added to simulate the various attacks that could be encountered in reality, such as the Cropout, Dropout, Gaussian blur, Gaussian noise, Resize, and JPEG Compression. A text-sensitive loss function is designed to limit the embedding modification on characters. An embedding strength adjustment strategy is proposed to improve the quality of watermarked image with little loss of extraction accuracy. Experimental results show that the proposed document image watermarking technology outperforms three state-of-the-arts in terms of the robustness and image quality.
\end{abstract}


\section{Introduction}\label{sec:introduction}

\begin{figure}[htbp]
		\includegraphics[width=\linewidth]{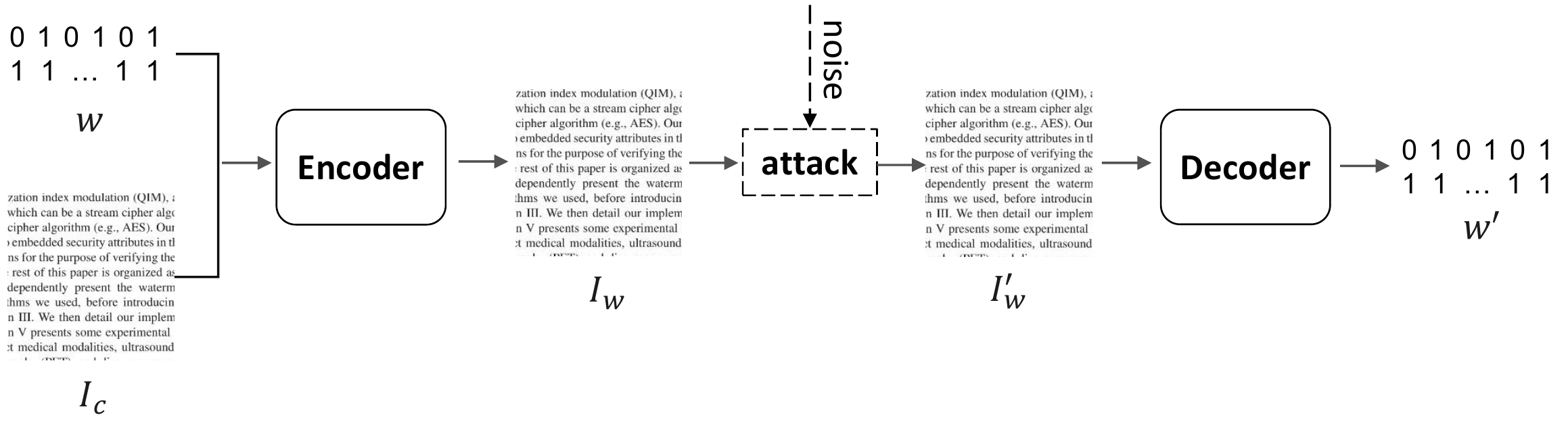}
		\caption{ A framework of the document image watermarking.}
		\label{fig:watermarking_framework}
	\end{figure}

 The development of computer multimedia technology provides great conveniences for e-commerce and e-government, in which many valuable documents, such as administrative documents, certificates, medical cases and transaction certificates, are scanned to be the document images in digital format. The document images can be stored and exchanged efficiently but can also be illegally copied and stolen by unauthorized persons. In addition to the encryption technologies for the confidentiality, watermarking techniques are generally designed to protect the copyright of image \cite{cox2007digital}. As shown in Fig.~\ref{fig:watermarking_framework}, the identity information of data owner can be embedded as the watermark $w$ into a cover document image $I_{c}$, generating a watermarked version denoted as $I_{w}$. During the storage and transmission, the watermarked image could suffer various attacks such as compressing, noising, blurring, resizing, and so on. If needed, the watermark should be extracted from the attacked version $I'_w$ to identify the copyright.

 Many watermarking technologies have been designed for the natural image, but document image watermarking schemes are also worth paying attention to. The existing document image watermarking technologies can be divided into two categories: the structure-based and the image-based methods \cite{fang2019camera}. The structure-based methods perform the embedding process by fine-tuning the text structure of the document, such as line shifting, word shifting, and character feature encoding \cite{brassil1994electronic, brassil1999copyright, amano1999feature, huang2001interword, kim2003text, tan2012printscan}. These approaches have good robustness but low embedding capacity \cite{kamaruddin2018review}. The image-based methods process the document image as a whole to embed the watermark. The embedding can be conducted in both the spatial domain \cite{kim2004watermarking, loc2018document} and transform domain \cite{lu2002watermark, rosiyadi2011copyright, chetanK2015efficient, al2017copyright, dang2019blind}, like the technologies for natural images. The image-based methods generally hold high embedding capacity, but not good in terms of imperceptibility and robustness.

 In recent few years, the Deep Neural Network (DNN) has been used to construct the watermarking schemes for natural images \cite{mun2019finding, ahmadi2020redmark, zhong2020automated, zhu2018hidden, liu2019novel, luo2020distortion}. These methods utilize the powerful fitting ability of DNN to automatically learn the natural image watermarking algorithms, getting the improved imperceptibility and robustness. However, these methods cannot be applied to document images without adjustments due to the apparent difference between the document and natural images. In this paper, we propose a robust document image watermarking scheme by using DNN. The main contributions can be concluded as follows: 
 
 \begin{itemize} 
	\item To the best of our knowledge, the proposed watermarking method is the first DNN-based one for document images. A text-sensitive loss function is designed to decrease the modification on text characters. The noise layer is constructed to simulate various attacks to improve the robustness. The watermark expansion strategy also helps to increase the robustness.
	
	\item It is found that the visual effect of the watermarked image is unsatisfying although the PSNR and SSIM seem good enough, which can be attributed to the clear background of the document image. Accordingly, we proposed an embedding strength adjustment strategy to increase the image quality with little loss of extraction accuracy.
	
	\item Due to the lack of ready-made document image dataset, we construct two large-scale document image datasets for the DNN training. One is named DocImgEN which includes 230,000 training, 10,000 validation and 10,000 testing document images with English sentences. The other is named DocImgCN including 230,000 training, 10,000 validation and 10,000 testing document images with Chinese sentences.
 \end{itemize}

 The rest of the paper is organized as follows. The related works are presented in Section~\ref{sec:releted_work}. The proposed scheme is described in Section~\ref{sec:proposed_scheme}. The experiment results and analysis are presented in Section~\ref{sec:experiments}. Finally, Section~\ref{sec:conclusion} draws the conclusion of our work.

 \section{Related works}\label{sec:releted_work}
 In this section, we firstly introduce the watermarking technologies for document images. Then several DNN-based watermarking technologies are discussed as we will use DNN to construct our method.

\subsection{Watermarking methods for document images}
 The existing document image watermarking methods can be divided into two categories: structure-based and image-based ones.

 \textbf{The structure-based methods.} This kind of methods embed watermarks into document images by exploiting the specific structure in text documents. Brassil \emph{et al.} are the first to study document image watermarking \cite{brassil1994electronic, brassil1999copyright}. Three technologies, i.e., line-shift coding, word-shift coding, and character coding, are proposed and discussed. The line-shift coding and word-shift coding in \cite{brassil1994electronic, brassil1999copyright} require the original unmarked document image for watermark extraction and get the low embedding capacity. The character coding also requires the original document for extraction and the watermark can be easily affected by local noise since it marks on local features \cite{huang2001interword}. Accordingly, Huang and Yan \cite{huang2001interword} proposed a document image watermarking method to achieve a blind extraction. The watermark is associated to a sine wave with the specific phase and frequency. The average inter-word spaces of lines in document are adjusted according to the sine wave. This method supports blind watermark extraction but also have low embedding capacity. Kim \emph{et al.} \cite{kim2003text} also proposed a document image watermarking method by adjusting the inter-word spaces. Firstly, the words in the document are categorized into different classes according to width of its adjacent words. Then, the segment, consisting of several adjacent words, can be also classified according to the class of the words in it. Finally, the watermark bits are embedded by modifying statistics of inter-word spaces in segment classes. This method embeds the same watermark bits in each class of segments and is robust even though some words or segments are missed. Amano and Misaki \cite{amano1999feature} proposed a document image watermarking method by changing the width of strokes. Specifically, a text area, such as a line of words, is divided into two separate parts. Then, according to the watermarking bit, the character strokes in one part are changed to be fatter and that in the other part are changed to be thinner. During the watermark extraction, the average width of the strokes in two parts are compared to figure out the bit. Tan \emph{et al.} \cite{tan2012printscan} also proposed a document image watermarking method based on strokes of Chinese characters. The watermark bits are embedded by modulating the direction of strokes, and the shuffling is used to balance embedding payload.

 \textbf{The image-based approachs.} This kind of methods process the document image as a whole during the watermarking. Kim \emph{et al.} \cite{kim2004watermarking} applied the Sobel edge operator to generate the edge direction histogram. Authors revealed that the normalized edge direction histograms generated from the document image blocks in the same language are quite constant. Thus, the document image can be divided into blocks. Some of blocks are chosen as the reference blocks while the others can be slightly modified to adjust its edge direction histogram according to the watermark bits. Loc \emph{et al.} \cite{loc2018document} stated that the layout of a document could be quite complex and proposed to divide the document image using full convolution networks. Then, watermark bits are embedded into the appropriate segments. Specifically, each appropriate image segment is divided into blocks. If all the pixel values in a block are larger than a threshold, the block will be further divided into two parts. The watermark bits are embedded by adjusting the average of the pixel values in two parts. Lu \emph{et al.} \cite{lu2002watermark} proposed a watermarking technology for the binary image in discrete cosine transform (DCT) domain. For the satisfying imperceptibility and robustness to binarization, the binary images are firstly blurred to be the gray ones by Gaussian filter. Then, the gray image is divided into $8\times 8$ blocks and the non-uniform blocks are transformed into DCT domain. The DC component is modified to carry the watermark bits. Finally, the whole gray image is binarized back. This method is proved to be robust to print and scan operations but the watermark extraction needs the participation of the original image. Horng \emph{et al.} \cite{rosiyadi2011copyright} also proposed a document image watermarking method in DCT domain. Firstly, the image is divided into $8\times 8$ blocks and then transformed by DCT. Next, the $8\times 8$ blocks are decomposed by singular value decomposition (SVD). Then the singular values are adjusted to embed the watermark bits. Chetan and Nirmala \cite{chetanK2015efficient} proposed to watermark document images in Discrete Wavelet Transform (DWT) domain. The document image is divided into segments and the non-empty segments are transformed by Level-2 DWT. Then the watermark bits are embedded in the LL2 sub-bands. Al-Haj and Barouqa \cite{al2017copyright} further proposed a watermarking method in DWT domain by using SVD. Similar to \cite{rosiyadi2011copyright}, the selected sub-band is decomposed by SVD and the singular values are adjusted to embed the watermark bits. Dang \emph{et al.} \cite{dang2019blind} proposed a watermarking scheme to embed a QR Code into the HH2 sub-band of the DWT domain. The coding strategy of QR Code can provide extra robustness to the noise. 
 
 Generally, the structure-based document image watermarking technologies have good robustness but low capacity. On the contrary, the image-based ones hold higher capacity but are not so robust to the noises.

\subsection{DNN-based watermarking methods for natural images}
 
 The Deep Neural Network designed for watermarking generally includes an \textit{encoder} and a \textit{decoder} for watermark embedding and extraction, respectively. Besides, the loss functions related to the image quality and watermark error are designed to train the encoder and decoder.
 
 Mun \emph{et al.} \cite{mun2019finding} proposed a robust watermarking method using DNN. The cover images are divided into non-overlapping blocks, and the encoder and decoder are designed to embed and extract the watermark in each block. To enhance the robustness, noises are added into the watermarked images which are then inputted into the decoder for reinforcement training. Considering that the watermarking in transform domain is more secure and robust against attacks, Ahmadi \emph{et al.} \cite{ahmadi2020redmark} proposed an embedding network structure including two transform layers. One layer is employed to transform the cover image into transform domain before the encoder while the other is used to transform the concatenation of cover and watermark back. Zhong \emph{et al.} \cite{zhong2020automated} proposed to add an Invariance Layer (\textbf{IL}) between the encoder and decoder. A regularization term is employed in \textbf{IL} to preserve useful information related to the watermark, while removing all other noise and irrelevant information. Zhu \emph{et al.} \cite{zhu2018hidden} proposed a noise layer to simulate various kinds of attacks in reality. During the training stage, the watermarked images from the encoder are input into the noise layer with random parameters. Then, the noised images are fed into the decoder for watermark extraction. Liu \emph{et al.} \cite{liu2019novel} stated that the one-stage end-to-end training can converge slowly and receive low-quality watermarked images due to the noise attack, and thus proposed a two-stage deep learning watermarking scheme. The first stage conducts a noise-free end-to-end training and the second refines the decoder with noise attacks. Luo \emph{et al.} \cite{luo2020distortion} stated that noisy layer in \cite{zhu2018hidden, liu2019novel} were the differentiable models and could generalize poorly to unknown distortions. The authors added a generative adversarial network between the encoder and decoder to generate distortion. In addition, channel coding is designed to add redundancy to the watermark, which also increases the robustness.

\begin{figure*}[ht]
	\centering
	\includegraphics[width=\linewidth]{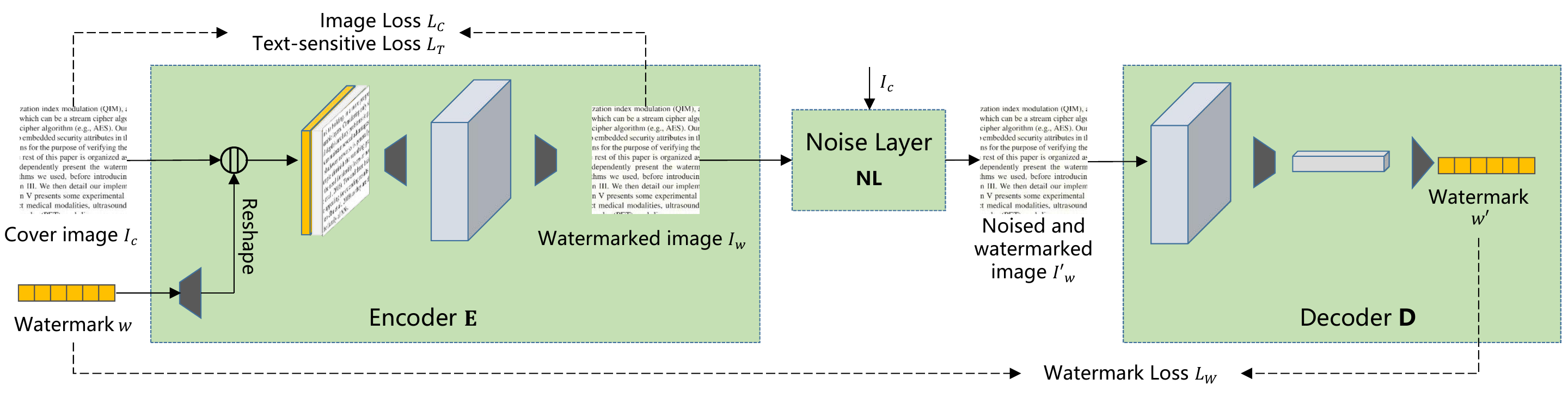}
	\caption{ The overview of the proposed scheme. In the training stage, the watermark $w$ is expanded, reshaped, and concatenated with the cover image $I_c$.  Then, combination of $I_c$ and $w$ are encoded to generate the watermarked $I_w$. Next, $I_w$ is input into the noise layer which simulates the attacks on input, generating the noised and watermarked image $I'_w$. Finally, $I'_w$ is input into the Decoder \textbf{D} to extract the watermark $w'$ which could be similar or the same to $w$. After training, the Encoder \textbf{E} is used to embed the watermark and the Decoder \textbf{D} is used for watermark extraction.}		
	\label{fig:overview}
\end{figure*}

\section{The proposed document image watermarking scheme}\label{sec:proposed_scheme}
 In this section, we firstly give an overview of our scheme. Then, the encoder, noise layer, decoder, and loss function are specified. 

\subsection{The overview of the proposed scheme}\label{sec:overview_proposed_scheme}
 Inspired by the DNN-based watermarking methods for natural images \cite{mun2019finding, ahmadi2020redmark, zhong2020automated, zhu2018hidden, liu2019novel, luo2020distortion}, we propose an end-to-end watermarking scheme for document images. As illustrated in Fig.~\ref{fig:overview}, the Encoder \textbf{E} and Decoder \textbf{D} are constructed to embed and extract watermarks, a Noise Layer (\textbf{NL}) is designed to simulate the possible distortion in reality, and the loss functions are calculated to optimize the encoder and decoder by considering the particularity of document images.

\subsection{Encoder}\label{sec:encoder}

\begin{figure*}[htbp]
	\centering
	\includegraphics[width=\linewidth]{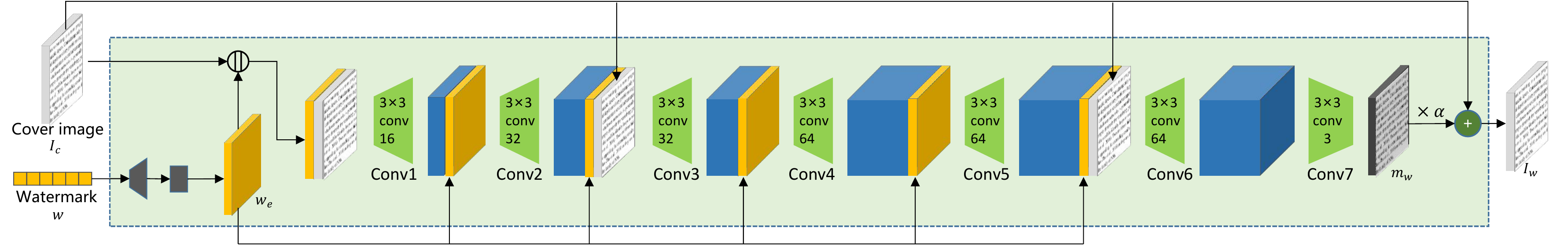}
	\caption{
		The structure of the encoder \textbf{E}.}
	\label{fig:encoder}
\end{figure*}

 The Encoder \textbf{E} is a network trained to embed the watermark into cover document image. Firstly, the watermark $w$ is expanded to bring in redundancy. Then the expanded watermark and cover are encoded together, generating a watermark mask $m_w$. Finally, the watermark mask is added to the cover to generate the watermarked image $I_w$. 
 
 \textbf{Watermark expansion}. In our scheme, the watermark $w$ can be an arbitrary string of binary bits. Before encoded with the cover, the watermark $w$ is expanded to bring in redundancy through a fully connected layer, which helps to enhance the robustness to noise attacks. Then the extended watermark is reshaped and upsampled to be a three-dimensional tensor $w_e$ with the same size as $I_c$.
 
 \textbf{Encoding}. The expanded watermark $w_e$ and the cover document image $I_c$ are concatenated and encoded through convolutional layer Conv1-7 as illustrated in Fig.~\ref{fig:encoder}, generating a watermark mask $m_w$. The DNN-based watermarking method is essentially using the convolutional maps to co-encode with the watermarks \cite{liu2019novel}. Thus, it is better to learn the watermarking mode by using convolutional maps of different levels. Accordingly, as illustrated in Fig.~\ref{fig:encoder}, the expanded watermark $w_e$ is concatenated to the output of Conv1-5, and the cover document image $I_c$ is concatenated to the output of Conv2 and Conv5.
 
 \textbf{Addition}. The convolutional layers output a three-channel watermark mask $m_w$ which is finally added to the cover as follow,
 
 \begin{equation} 
 \label{equ:addition}
 	I_w = I_c + \alpha \cdot m_w,
 \end{equation}
 
 \noindent where $\alpha$ is an embedding strength factor.
 
 \subsection{Noise layer}\label{sec:noise_layer}

\begin{figure}[htbp]
	\includegraphics[width=\linewidth]{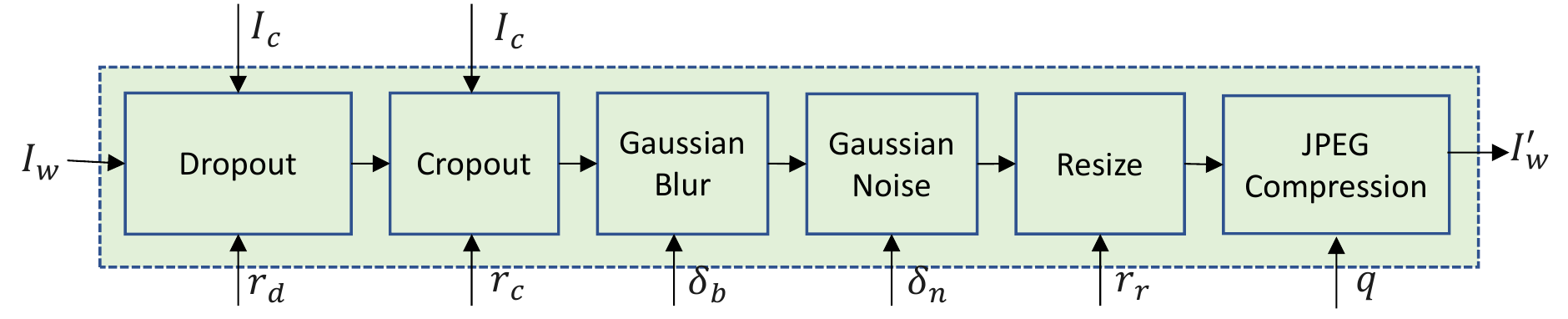}
	\caption{
		The structure of Noise Layer \textbf{NL}.}
	\label{fig:noise_layer}
\end{figure}

 The watermarked image can suffer various distortions during the storage and transmission. The watermarking scheme cannot be robust to these attacks without appropriate training. Accordingly, a Noise Layer (\textbf{NL}) is added between the encoder and decoder in the training stage to bring in distortions, such as Dropout, Cropout, Gaussian Blur, Gaussian Noise, Resize, and JPEG Compression, as illustrated in Fig.~\ref{fig:noise_layer}. The attacks in reality are expected to be the individual or combination of these distortions.

	\begin{itemize} 
		\item \textit{Dropout}. A part of pixels in $I_w$ is randomly selected and replaced by the pixels in $I_c$ at the corresponding positions. The ratio of selected pixels is denoted as $r_d$ and we set $r_d \le 10\%$.
		\item \textit{Cropout}. The pixels in a randomly chosen region of $I_w$ keep unchanged, and the rest is replaced by the corresponding part of $I_c$. The ratio of the unchosen region to the whole $I_w$ is denoted as $r_c$ and we set $r_c \le 10\%$. 
		\item \textit{Gaussian Blur}: A Gaussian blur is conducted on $I_w$ with the window size of $ \delta_b \times \delta_b$ and a random standard deviation from $[1.0,3.0]$.
		\item \textit{Gaussian noise}. Gaussian noise with a random standard deviation $\delta_n \in [0,0.02) $ is generated and added to $I_w$.
		\item \textit{Resize}: The size of watermarked image $I_w$ is reduced and then amplified to the original size. The reduction ratio $r_r$ is randomly chosen from $[0,50\%]$.
		\item \textit{JPEG Compression}: JPEG Compression includes several steps such as color space conversion, block splitting, discrete cosine transformation, quantization, and so on. Among these steps, the quantization cannot be directly incorporated into the training network as it is not differentiable. Thus, we use the method in \cite{shin2017jpeg, luo2021rate} to approximate the effect of quantization step as,
		
		\begin{scriptsize} \begin{equation}
				 d_i \leftarrow \lfloor {( \frac{d_i}{M_{\#}^{i} \times q_f} + (\lfloor \frac{d_i}{M_{\#}^{i} \times q_f} - \frac{d_i}{M_{\#}^{i} \times q_f} \rceil )^3) \cdot {(M_{\#}^{i} \times q_f)}}\rceil,
		\end{equation} \end{scriptsize}

		\noindent where $\lfloor \cdot \rceil$ represents the rounding to the nearest integer, $d_i$ means the DCT coefficients, $M_{\#}, \#\in \{Y,C\}$ are the standard quantization table for $Y$ component and $CbCr$ components, and $q_f$ is calculated as follow,
		
		\begin{equation}
			q_f =\begin{cases} 
			50.0/q + 0.0001, & q<50, \\ 
			2.0 - 0.02q + 0.0001, & q \ge 50,
			\end{cases} 
		\end{equation}
	
		\noindent where $q$ denotes the JPEG quality factor and uniformly selected within $[50,100)$ during the training process. Please note that, the real JPEG compression can be directly utilized in the testing process.
	\end{itemize}

\begin{figure}[htbp]
 	\includegraphics[width=\linewidth]{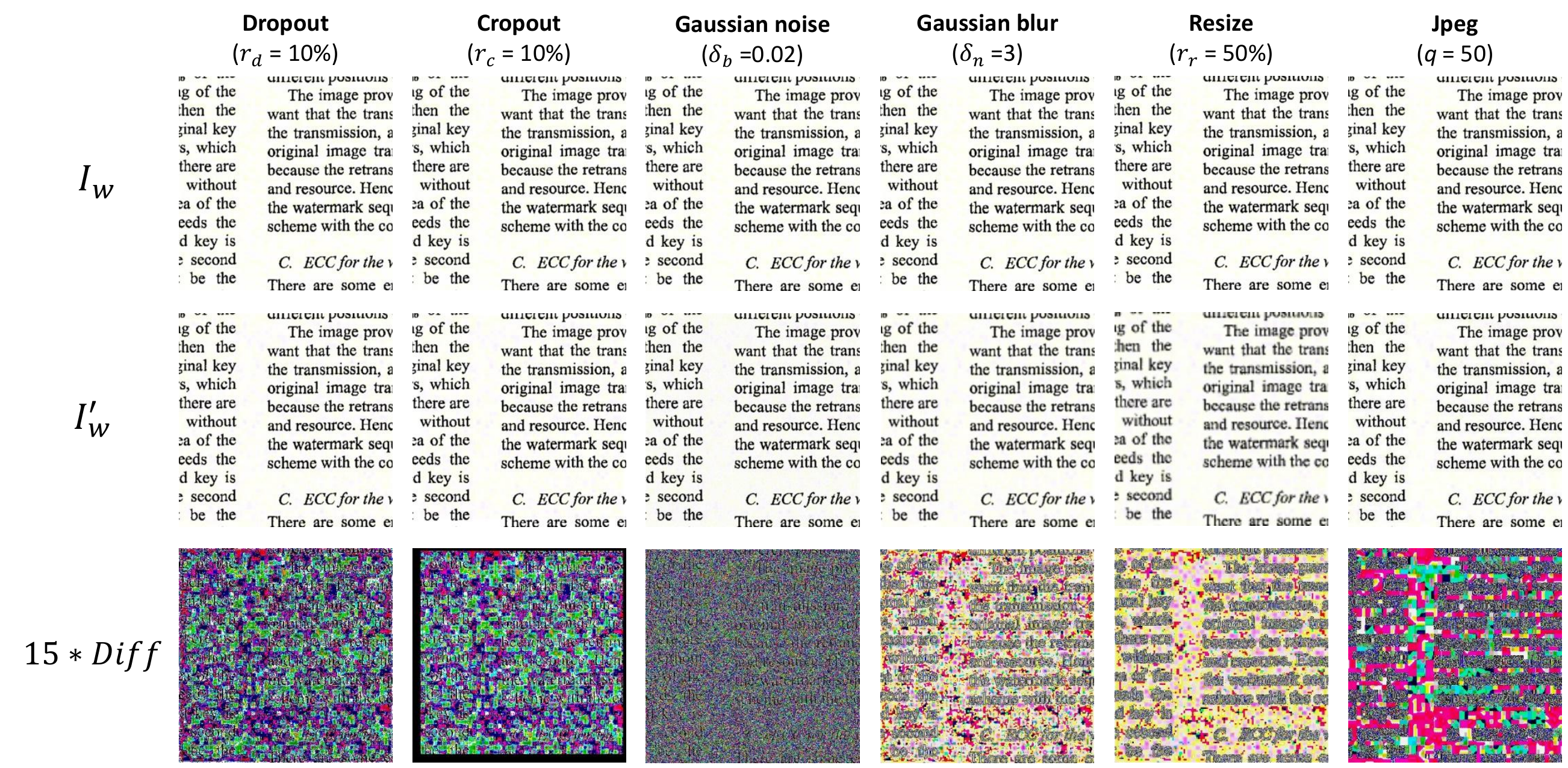}
 	\caption{
 		The visual examples of distortions.}
 	\label{fig:noise_layer_samples}
 \end{figure}

 The distortions above are illustrated in Fig.~\ref{fig:noise_layer_samples}. 
 
\subsection{Decoder}\label{sec:decoder}

\begin{figure*}[htbp]
	\includegraphics[width=\linewidth]{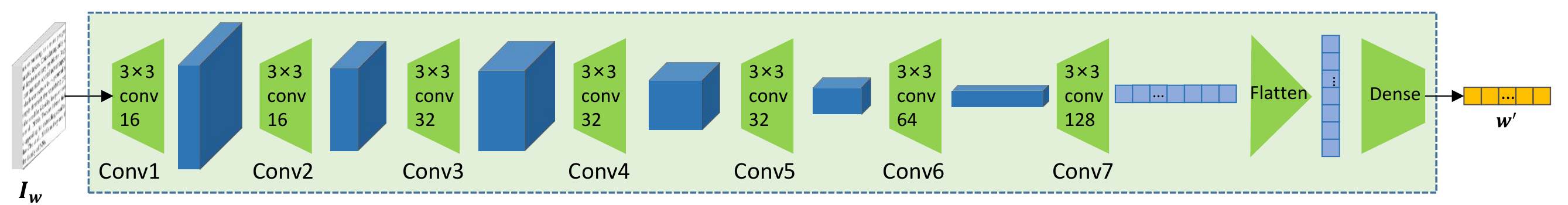}
	\caption{
		The structure of the Decoder \textbf{D}. }
	\label{fig:decoder}
\end{figure*}

 The Decoder \textbf{D} is trained to extract the watermark ($w'$) from the watermarked image. In the training stage, the input of \textbf{D} is the noisy watermarked image $I'_w$. Inspired by AlexNet \cite{krizhevsky2012imagenet}, the decoder consists of seven convolutional layers, a flatten layer, and a dense layer, as illustrated in Fig.~\ref{fig:decoder}. The stride size of the convolutional layer Conv1 and Conv3 is set to be the default value 1, while the others are set to 2 to speed up the training. After Conv7, a flatten operation is performed on the remaining neurons, and then a dense layer is used to output an 1-dimensional tensor which has the same size of $w$. Sigmoid function is used to produce a output in the last layer. 

\subsection{Loss function}\label{sec:loss_function}

 The loss function in our scheme consists of three parts: image loss, text-sensitive loss, and watermark loss.
 
 \textbf{Image loss}. Image loss is designed to keep the watermarked image $I_w$ similar to the cover $I_c$. We consider the image in YUV color space and try to make little change on the Y component as the human eyes are more sensitive to it. In addition, the large modification on a single pixel can be more destructive to visual effect. Accordingly, we restrain the big change on pixels. The image loss is designed as 

	\begin{equation}
	\begin{split}
		L_{I} 	&= MSE(I_{w}^{Y},I_{c}^{Y}) \times s_{Y} \\
		 		&+ MSE(I_{w}^{U},I_{c}^{U}) \times s_{U} \\
		     	&+ MSE(I_{w}^{V},I_{c}^{V}) \times s_{V},
	\end{split} 
	\end{equation}
 \noindent where $MSE$ denotes the mean squared error, and $s_{Y}$, $s_{U}$ and $s_{V}$ are the weights for YUV channels. 
 
 \textbf{Text-sensitive loss}. Readers will pay more attention to the characters during reading. The modification on characters can be more conspicuous \cite{zhao2016loss, fang2019camera}. Thus, the text-sensitive loss is designed to restrain the modification on characters as follows,
	\begin{equation}
 	\begin{split}
		L_{T} 	&= \mid I_{w}^R - I_{c}^R \mid \cdot \tilde{I_{c}^R} \times s_{R}\\ 
			&+ \mid I_{w}^G - I_{c}^G \mid \cdot \tilde{I_{c}^G} \times s_{G}\\
			&+ \mid I_{w}^B - I_{c}^B \mid \cdot \tilde{I_{c}^B} \times s_{B},
	\end{split} 
	\end{equation}

 \noindent where $\tilde{I_{c}^*}=\frac{255-I_{c}^*}{255}$ assigns larger punishment to the dark points, and $s_{*}$ denotes the weights for different color components, $* \in \{R,G,B\}$.

 \textbf{Watermark loss}. Watermark loss is designed to keep the extracted watermark as similar as the original. Binary cross entropy function is used for it as 
 
	 \begin{equation} 
	 	L_{W} = -\sum_{i=1}^{N}{(w_i \cdot \log(w'_i) + (1 - w_i) \cdot \log(1 - w'_i))}, 
	 \end{equation} 
	 
 \noindent where $w_i$ refers to the original watermark, $w'_i$ denotes the extracted watermark and $N$ is the bit number of the watermark.

Finally, the total training loss is calculated as
	\begin{equation} 
		L_{total} = \lambda_{I}L_{I} + \lambda_{T}L_{T} + \lambda_{W}L_{W}. 
	\end{equation}
 \noindent where $\lambda_{I}$, $\lambda_{T}$, and $\lambda_{W}$ are weight factors.

\section{Experimental results and analysis}\label{sec:experiments}

 This section presents our self-made datasets, implementation details, and experimental results. Besides, we discuss an embedding strength adjustment strategy which increases the image quality without much loss of extraction accuracy.

\subsection{The construction of document image datasets}\label{sec:document_image_data_set_construction}

\begin{figure}[htbp]
	\includegraphics[width=\linewidth]{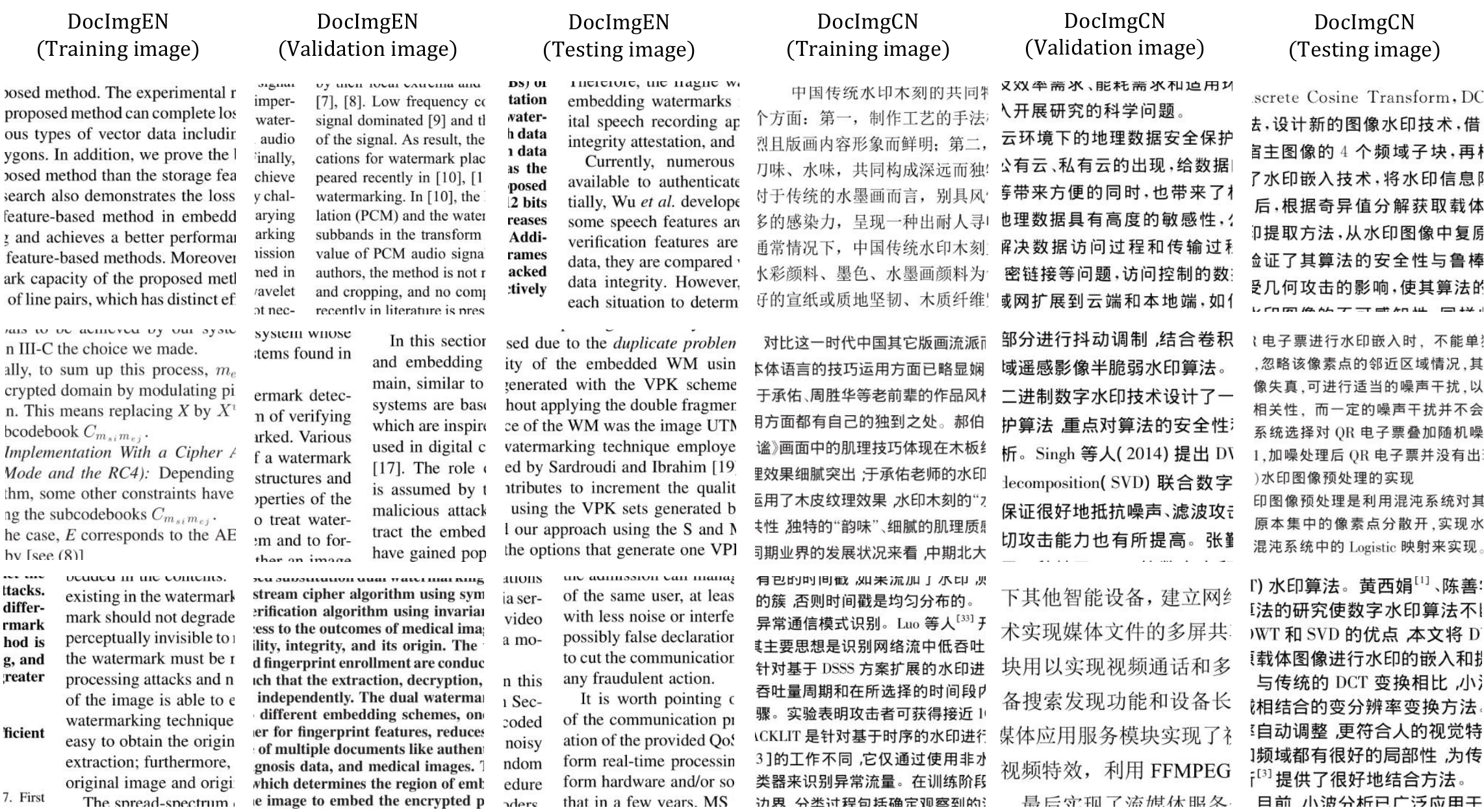}
	\caption{The examples of the document images in DocImgEN and DocImgCN datasets.}
	\label{fig:Dataset_examples}
\end{figure}

 Unlike the traditional document watermarking technologies, the DNN-based method needs to be trained with the large-scale training dataset. However, we regrettably find that there are currently no such datasets. Accordingly, we construct two large-scale document image datasets: DocImgEN and DocImgCN, and the examples are shown in Fig.~\ref{fig:Dataset_examples}. Please note that, the datasets can be downloaded for research.

 \textbf{DocImgEN}. We download a batch of PDFs from IEEE Xplore database \cite{ieee2021database} and convert the PDF pages to PNG  images. Then, the image blocks with the size $400 \times 400$ are cropped out as the document images. DocImgEN includes 230,000 training, 10,000 validation and 10,000 testing document images with English words. 

 \textbf{DocImgCN}. We download PDFs from China National Knowledge Infrastructure (CNKI) \cite{cnki2021database} to prepare document images with Chinese characters. Similarly, DocImgCN also includes 230,000 training, 10,000 validation and 10,000 testing document images.

\subsection{Implementation Details}\label{sec:implentation_details}
 
 We implement and test our scheme on TensorFlow with a GPU: NVADIA GTX 1080Ti. The parameters in our scheme are summarized in Table~\ref{Table:parameters}. The first 100,000 training images in DocImgEN and DocImgCN are utilized to train the corresponding models. Four images and four random bit strings assemble a training batch. The learning rate is set to be 0.0001 and Adam optimizer \cite{kingma2014adam} is employed to optimize the models. All the testing images in DocImgEN and DocImgCN are used to verify the performance.

 During the training stage, it is found that sometimes the decoder failed to achieve a satisfying extraction accuracy even with a large number of training iterations. Thus, we froze the encoder at the first 3000 iterations, just training the decoder and noisy layer. In addition, to make the decoder gradually adapt to the noise distortion, the parameters $\lambda_{I}$, $\lambda_{T}$, and $\lambda_{W}$ are set to be 0 at the beginning, and increase linearly to 1.5, 1.5, and 2.0 at the first 15,000 iterations.

	\begin{table}[ht]
		\caption{The setting of parameters in the training stage.}
		\centering
		\begin{tabular}{c|ccc}
			\hline \hline
			Parameter 			& $r_d$       	& $r_c$          	& $r_r$ 		\\ 
			Range				& $(0,10\%]$   	& $(0,10\%]$      	& $[0,50\%]$ 	\\ \hline
			Parameter         	& $\delta_b$    & $\delta_n$        & $q$ 			\\
			Range or value 		& $3$           & $(0,0.02]$       	& $[50,100)$ 	\\ \hline
			Parameter         	& $s_Y$         & $s_U$        		& $s_V$ 		\\
			Value  				& 100			& 1					& 1				\\ \hline
			Parameter         	& $s_R$         & $s_G$         	& $s_B$ 		\\
			Value         		& 3         	& 6         		& 1 			\\ \hline 
			Parameter         	& $\lambda_I$   & $\lambda_T$       & $\lambda_W$ 	\\
			Value         		& 1.5         	& 1.5         		& 2.0 			\\ \hline \hline
		\end{tabular}
		\label{Table:parameters}
	\end{table}

\subsection{The robustness of the proposed scheme}\label{sec:robustness_of_proposed_scheme}
 In our scheme, the noise layer is an important strategy to improve the robustness of the watermark. The tests are conducted on DocImgEN and DocImgCN separately. The watermark length here is 100 bits. Please note that, as listed in Table~\ref{Table:parameters}, we perform the distortions with a relatively low intensity in noise layer during the training stage, so as to guarantee the quality of the watermarked image $I_w$. After training, the trained models are tested with higher intensity distortions to verify the robustness. 
 
\begin{figure}[htbp]
		\includegraphics[width=\linewidth]{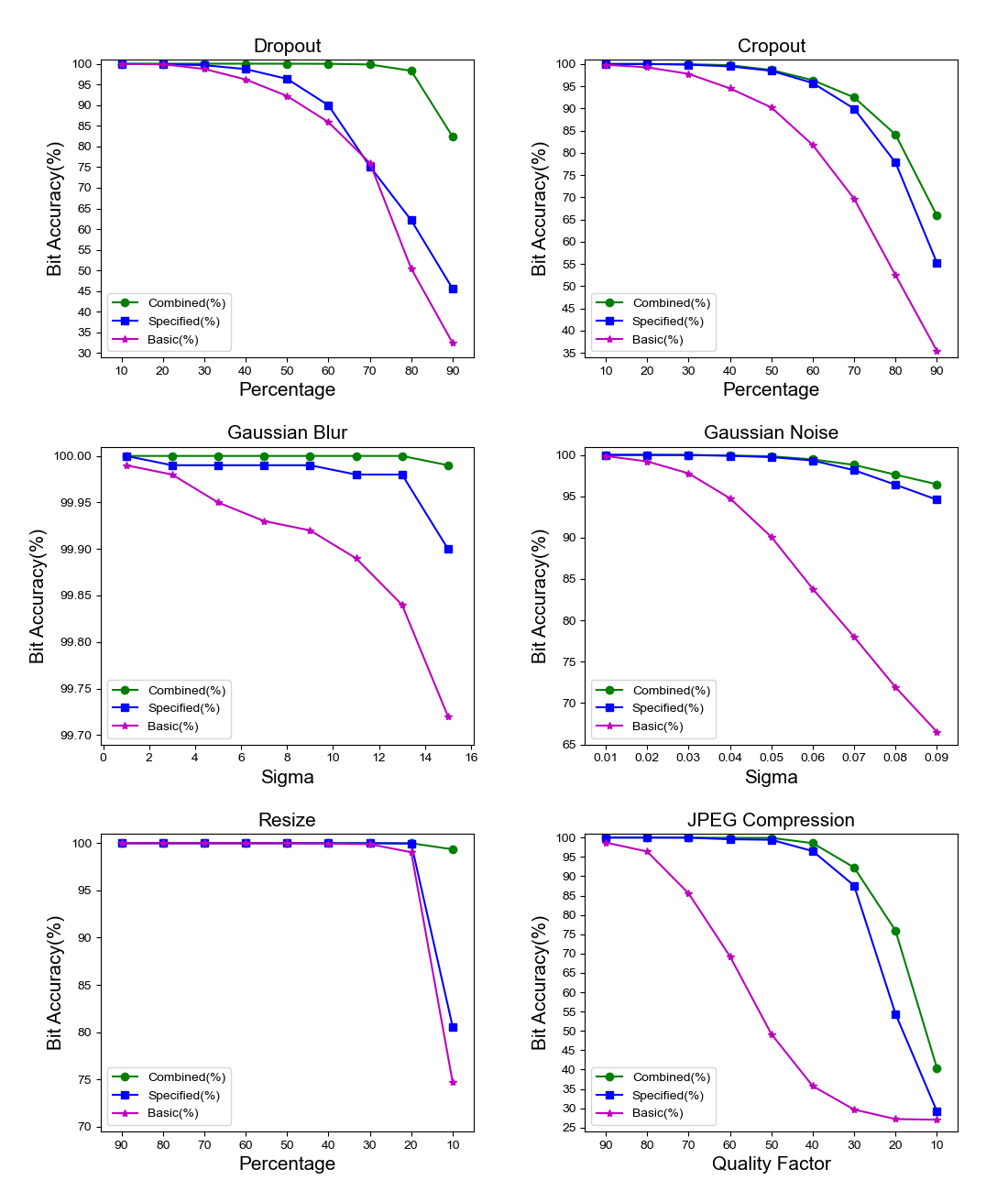}
		\caption{Bit accuracy of our schemes against six types of distortions in DocImgEN dataset.}
		\label{fig:DocImgEN_robust}
	\end{figure}

\begin{figure}[htbp]
		\includegraphics[width=\linewidth]{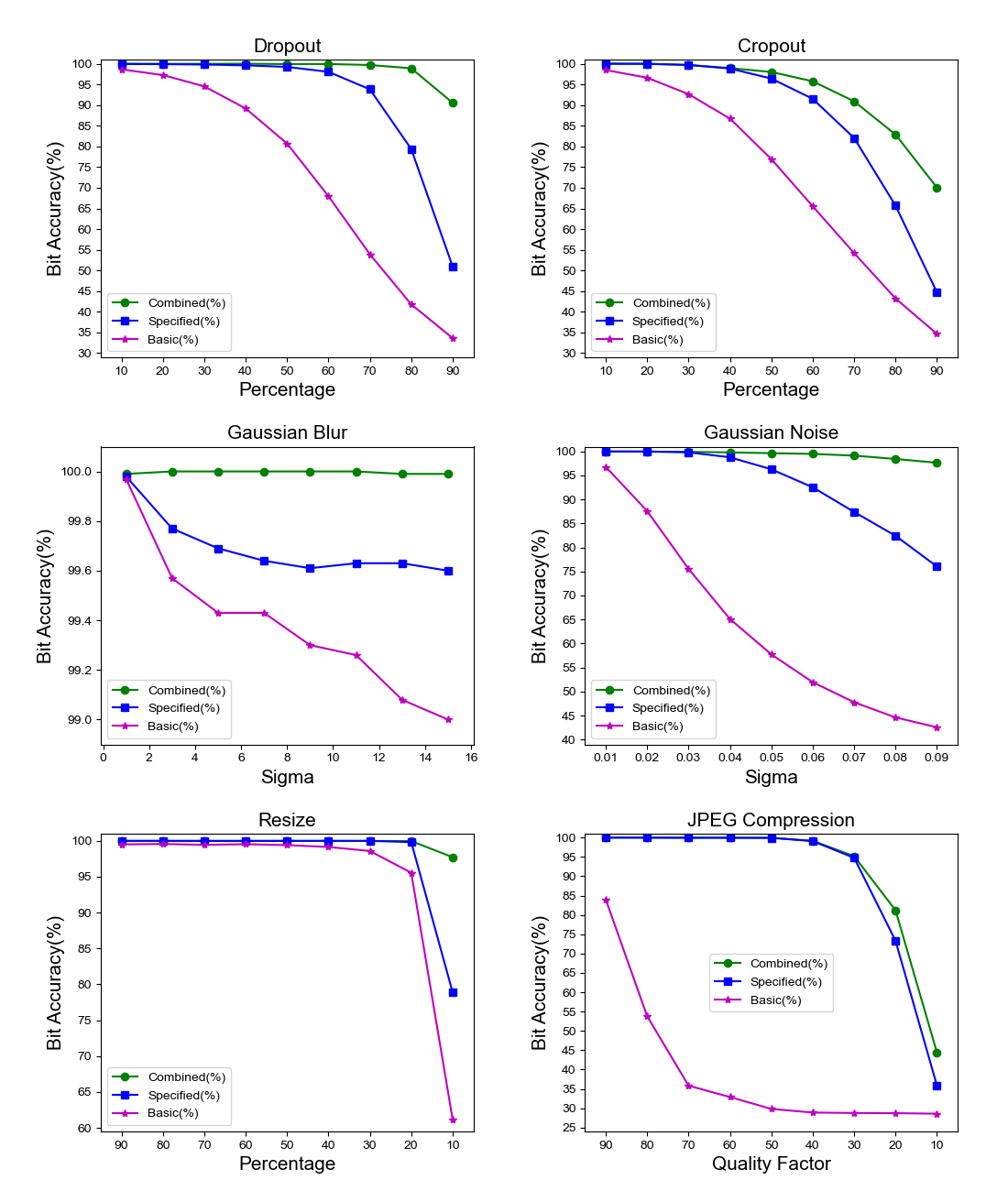}
		\caption{Bit accuracy of our schemes against six types of distortions in DocImgCN dataset.}
		\label{fig:DocImgCN_robust}
	\end{figure}

 Here we test our scheme in three different cases. At first, the encoder and decoder are trained without the noise layer. The generated model in this case is named as \textit{basic model}. Next, the encoder and decoder are trained with the noise layer that just considers a single distortion. Accordingly, six models are generated for six types of distortions per each dataset, respectively, and named as \textit{specified models}. Finally, the encoder and decoder are trained with the noise layer that considers all six distortions together. The resulting model is named as \textit{combined model}. The training iterations for the \textit{basic}, \textit{specified}, and \textit{combined} models are set to be 50,000, 60,000, and 80,000, respectively, which are large enough for adequate training. The results for DocImgEN and DocImgCN are shown in Fig.~\ref{fig:DocImgEN_robust} and \ref{fig:DocImgCN_robust}. 
 
 As shown in Fig.~\ref{fig:DocImgEN_robust} and \ref{fig:DocImgCN_robust}, although noise layer is not incorporated in training, the \textit{basic model} is still robust to the distortions to some extent, especially to Gaussian blur and resize. This robustness derives from the DNN structure in encoder and decoder. Next, the \textit{specified model} that considers a single distortion is tested by the identical distortion with various intensity. Substantial improvements are achieved. Most importantly, the \textit{combined model} gets better robustness than the \textit{specified models} for all six distortions, especially with the higher distortion intensity. It indicates that the encoder and decoder in our scheme are powerful enough to learn appropriate embedding strategies for resisting various distortions. Besides, the comprehensive consideration of various distortions can improve robustness to each single distortions. The results in following are generated by the \textit{combined model}. 
 
\subsection{The quality of the watermarked document image}\label{sec:watermarked_document_image_quality}

 The visual quality of the watermarked image is guaranteed by the image loss and text-sensitive loss. Two common measurements, i.e., Peak Signal to Noise Ratio (PSNR) and Structural Similarity Metric (SSIM) \cite{wang2004image}, are utilized to evaluate the similarity between the cover and watermarked images. Specifically, PSNR measures the similarity in pixel level while SSIM calculates similarity from the brightness, contrast, and structure. Considering that readers may be more sensitive to the characters while reading the documents, we designed the text-sensitive loss to ensure less modification on text pixels. Here we define the \textbf{C}hange Intensity \textbf{P}er Text-\textbf{P}ixel (CPP) to evaluate the modification on text pixels as follows,
 
	\begin{equation} 
	\begin{split}
		CPP &= \frac{\sum_{i=1}^{n_t}{\mid I_{c}^{tR}(i) - I_{w}^{tR}(i) \mid}}{n_t}\\ 
			&+ \frac{\sum_{i=1}^{n_t}{\mid I_{c}^{tG}(i) - I_{w}^{tG}(i) \mid}}{n_t}\\
			&+ \frac{\sum_{i=1}^{n_t}{\mid I_{c}^{tB}(i) - I_{w}^{tB}(i) \mid}}{n_t},
	\end{split} 
	\end{equation}

 \noindent where $I_{\#}^{t*}, \#\in {c,w}, *\in \{R,G,B\} $ refers to the set of text pixels in the cover and watermarked images, and $n_t$ denotes the total number of the text pixels. 
	
\begin{figure}[htbp]
		\includegraphics[width=\linewidth]{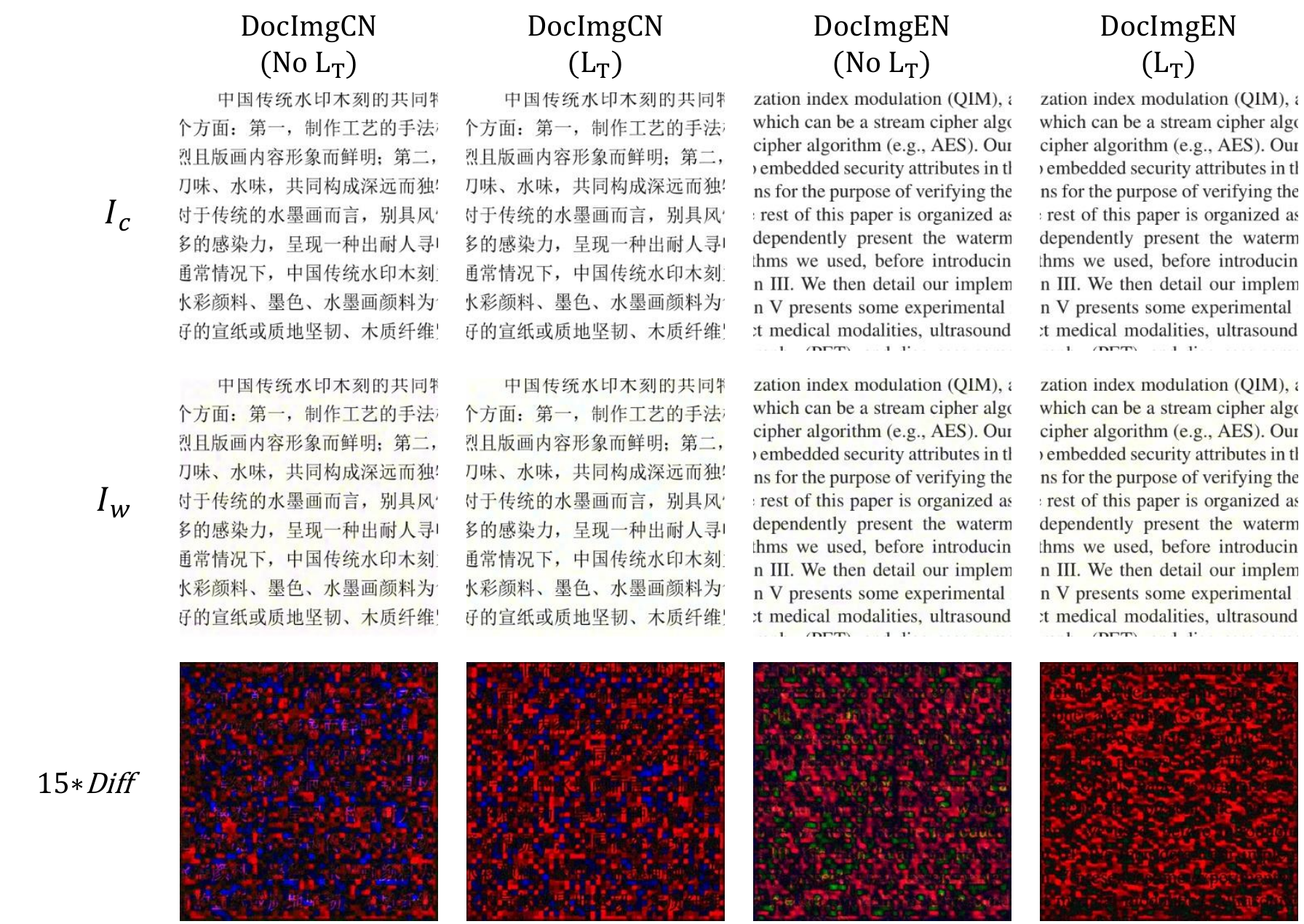}
		\caption{Example pairs of cover and watermarked images and their difference with the text-sensitive loss $L_{T}$ or not.}
		\label{fig:imagequality}
	\end{figure}

 During the testing of image quality, the watermark length is also set to be 100 bits. The PSNR, SSIM, and CPP values are listed in Table~\ref{Table:ImgQuality}, and some example pairs of cover and watermarked images are shown in Fig.~\ref{fig:imagequality}. The results are averaged from all testing images in DocImgEN and DocImgCN testsets. As listed in Table~\ref{Table:ImgQuality}, the incorporation of text-sensitive loss $L_T$ make little influence on PSNR and SSIM but decrease CPP by 57.52\% averagely. It indicates that $L_T$ decreases the modification on characters without much influence on image quality. 

	\begin{table}[ht]
		\caption{
			PSNR, SSIM, and CPP of our scheme with the text-sensitive loss $L_{T}$ or not.}
		\centering
		\begin{tabular}{l|ccc}
			\hline \hline
			\makecell[c]{Schemes}        	& PSNR (dB)      & SSIM           & CPP \\ \hline
			On DocImgCN without $L_{T}$     & 39.30          & \textbf{0.972} & 8.85 \\
			On DocImgCN with $L_{T}$        & \textbf{41.07} & 0.965          & \textbf{3.77} \\ \hline
			On DocImgEN without $L_{T}$     & 37.90          & \textbf{0.969} & 8.31 \\
			On DocImgEN	with $L_{T}$        & \textbf{40.10} & 0.962          & \textbf{3.52} \\ \hline \hline
		\end{tabular}
		\label{Table:ImgQuality}
	\end{table}

\subsection{Comparison with the state-of-the-arts}\label{sec:comparison_with_the_state_of_the_art}

 We compared our scheme with three DNN-based watermarking models, i.e., HiDDeN \cite{zhu2018hidden}, Liu \emph{et al.} \cite{liu2019novel} and Stegastamp \cite{tancik2020stegastamp}, in terms of robustness and image quality. As the three models are trained for nature images, we retrained these end-to-end models on DocImgEN and DocImgCN. The length of the watermark is 100 bits and the rest training parameters are set as that in original paper \cite{zhu2018hidden, liu2019novel, tancik2020stegastamp}. Please note that, the noise distortions added in the noise layer are the same as ours.

\begin{figure*}[htbp]
		\includegraphics[width=\linewidth]{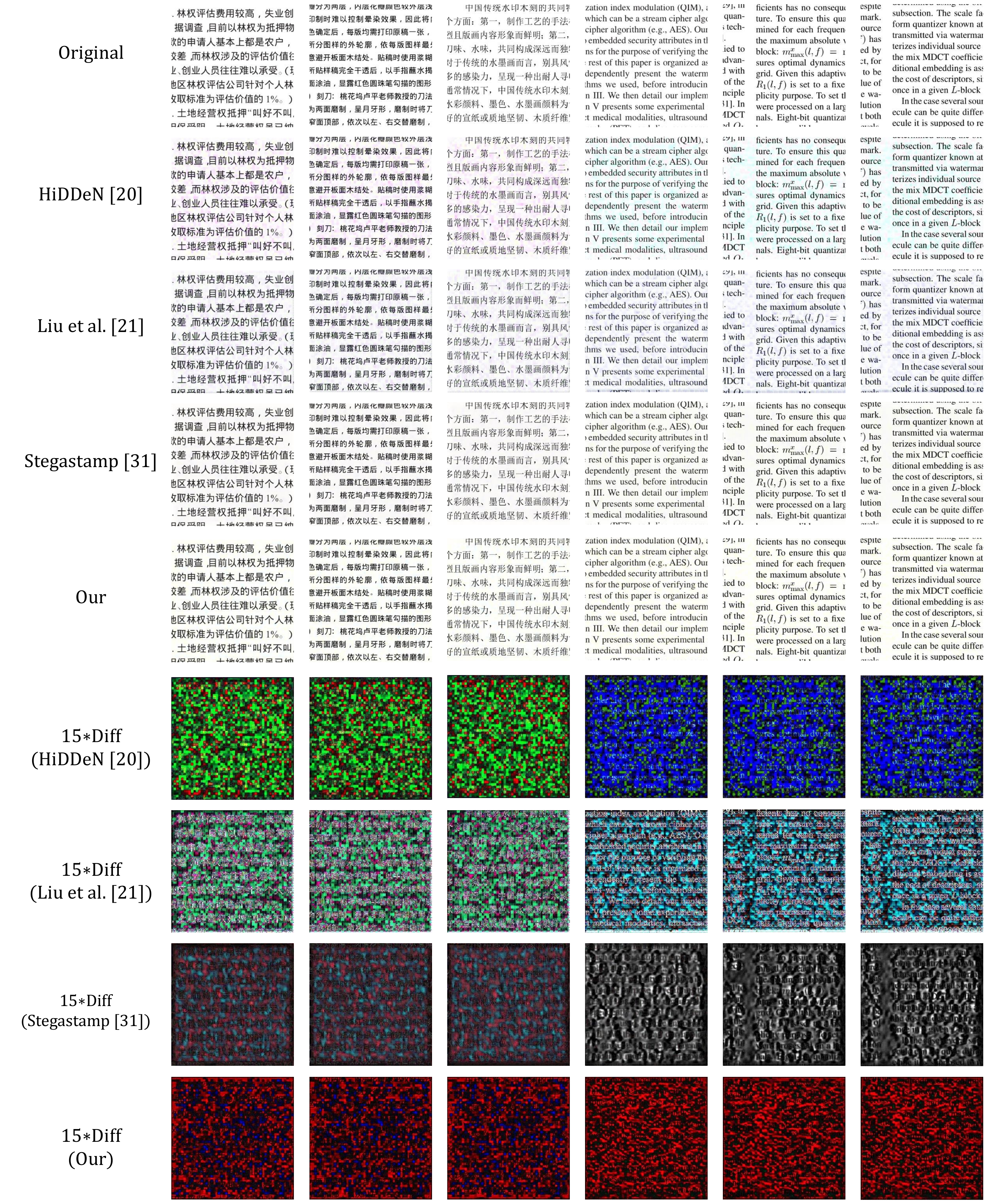}
		\caption{
			The examples of watermarked images from our scheme and state-of-the-arts.}
		\label{fig:QualityComparison}
	\end{figure*}

 The bit accuracy is calculated under different distortions and listed in Table~\ref{Table:RobustComparisonEN} and \ref{Table:RobustComparisonCN}. It shows that our scheme achieves the best robustness. The schemes in \cite{zhu2018hidden, liu2019novel, tancik2020stegastamp} are designed for nature images while our scheme is specially designed for the document image. In addition, some effective operations in \cite{zhu2018hidden, liu2019novel, tancik2020stegastamp} are kept in our model such as the noise layer, watermark expansion, and concatenation of watermark during the convolution process. As listed in Table~\ref{Table:QualityComparison}, our scheme also achieves the best PSNR and CPP while Stegastamp \cite{tancik2020stegastamp} holds better SSIM than ours. The perceptual loss \cite{zhang2018unreasonable} could be helpful for high SSIM. Finally, the watermarked images by \cite{zhu2018hidden, liu2019novel, tancik2020stegastamp} are illustrated in Fig.~\ref{fig:QualityComparison}. 

 \begin{table}[ht]
	\caption{
		The PSNR, SSIM, and CPP of schemes on the DocImgEN and DocImgCN.}
	\centering
	\begin{tabular}{l|ccc}
		\hline \hline
		\makecell[c]{Schemes}								& PSNR (dB)					& SSIM				& CPP \\ \hline
		HiDDeN \cite{zhu2018hidden} (On DocImgEN)			& 33.80						& 0.937				& 14.75 \\
		Liu \emph{et al.} \cite{liu2019novel} (On DocImgEN)	& 27.20						& 0.891				& 39.89 \\
		Stegastamp\cite{tancik2020stegastamp} (On DocImgEN)	& 34.79						& \textbf{0.977}	& 7.43 \\
		Ours (On DocImgEN)									& \textbf{40.10}			& 0.962				& \textbf{3.52} 	\\ \hline
		HiDDeN \cite{zhu2018hidden} (On DocImgCN)			& 33.70						& 0.939				& 12.47 \\
		Liu \emph{et al.} \cite{liu2019novel} (On DocImgCN)	& 28.20						& 0.901				& 32.15 \\
		Stegastamp\cite{tancik2020stegastamp} (On DocImgCN)	& 35.70						& \textbf{0.984}	& 8.35 \\
		Ours (On DocImgCN)									& \textbf{41.07}			& 0.965				& \textbf{3.77} \\ 
		\hline \hline
	\end{tabular}
	\label{Table:QualityComparison}
 \end{table}

	\begin{table*}[ht]
		\caption{Bit accuracy of schemes on the DocImgEN.}
		\centering
		\begin{tabular}{l|cccc}
			\hline \hline
			\makecell[c]{Attack Type}	& HiDDeN \cite{zhu2018hidden}	& Liu \emph{et al.} \cite{liu2019novel}	& Stegastamp \cite{tancik2020stegastamp}	& Ours \\ \hline
			Dropout ($r_d$=10\%)			 & 93.33	& 96.34 								& 99.96 								& \textbf{100} \\
			Dropout ($r_d$=30\%)			 & 84.76	& 86.25 								& 99.92 								& \textbf{100} \\
			Dropout ($r_d$=50\%)			 & 75.85	& 74.73 								& 99.73 								& \textbf{99.99} \\ \hline
			Cropout ($r_c$=10\%)			 & 93.06	& 82.26 								& 99.93 								& \textbf{100} \\
			Cropout ($r_c$=30\%)			 & 83.86	& 76.21 								& 99.83 								& \textbf{99.95} \\
			Cropout ($r_c$=50\%)			 & 82.16	& 75.47 								& \textbf{98.57} 						& 98.48 \\ \hline
			Gaussian Blur ($\delta_b$=3)	 & 95.13	& 98.88 								& 99.99 								& \textbf{100} \\
			Gaussian Blur ($\delta_b$=5)	 & 94.72	& 98.75 								& 99.99 								& \textbf{100} \\
			Gaussian Blur ($\delta_b$=7)	 & 94.23	& 98.57 								& 99.97 								& \textbf{100}\\ \hline
			Gaussian Noise ($\delta_n$=0.02) & 94.92	& 98.27 								& 99.84 								& \textbf{100} \\
			Gaussian Noise ($\delta_n$=0.03) & 93.31	& 96.88 								& 99.68 								& \textbf{100} \\
			Gaussian Noise ($\delta_n$=0.05) & 89.93	& 92.01 								& 98.11 								& \textbf{99.83} \\ \hline
			Resize ($r_r$=50\%)				 & 94.40	& 98.70 								& 99.99 								& \textbf{100} \\
			Resize ($r_r$=30\%)				 & 91.61	& 97.85 								& 99.91 								& \textbf{99.99} \\
			Resize ($r_r$=10\%)				 & 79.20	& 78.32 								& 78.96 								& \textbf{97.53} \\ \hline
			JPEG Compression ($q$=50)		 & 90.17	& 93.76 								& 99.68 								& \textbf{99.81} \\
			JPEG Compression ($q$=30)		 & 85.14	& 91.85 								& 92.91 								& \textbf{96.62} \\
			JPEG Compression ($q$=20)		 & 80.73	& 87.40 								& 86.29 								& \textbf{88.68} \\ 
			\hline \hline
		\end{tabular}
		\label{Table:RobustComparisonEN}
	\end{table*}

	\begin{table*}[ht]
		\caption{Bit accuracy of schemes on the DocImgCN.}
		\centering
		\begin{tabular}{l|cccc}
			\hline \hline
			\makecell[c]{Attack Type}	& HiDDeN \cite{zhu2018hidden}	& Liu \emph{et al.} \cite{liu2019novel}	& Stegastamp \cite{tancik2020stegastamp}	& Ours \\ \hline
			Dropout ($r_d$=10\%)			 & 98.20 	& \textbf{100} 							& 99.67 								& \textbf{100} \\
			Dropout ($r_d$=30\%)			 & 90.68 	& 99.92 								& 99.44 								& \textbf{99.96} \\
			Dropout ($r_d$=50\%)			 & 77.40 	& 94.18 								& 99.22 								& \textbf{99.90} \\ \hline
			Cropout ($r_c$=10\%)			 & 97.49 	& 99.55 								& 99.74 								& \textbf{99.99} \\
			Cropout ($r_c$=30\%)			 & 84.67 	& 82.89 								& 99.15 								& \textbf{99.68} \\
			Cropout ($r_c$=50\%)			 & 79.79 	& 77.01 								& \textbf{98.31} 						& 97.99 \\ \hline
			Gaussian Blur ($\delta_b$=3)	 & 99.13 	& \textbf{100} 							& 99.68 								& \textbf{100} \\
			Gaussian Blur ($\delta_b$=5)	 & 98.90 	& \textbf{100} 							& 99.66 								& \textbf{100} \\
			Gaussian Blur ($\delta_b$=7)	 & 98.69 	& \textbf{100} 							& 99.58 								& \textbf{100} \\ \hline
			Gaussian Noise ($\delta_n$=0.02) & 97.10 	& \textbf{99.99} 						& 99.45 								& 99.95 \\
			Gaussian Noise ($\delta_n$=0.03) & 94.10 	& 99.85 								& 99.35 								& \textbf{99.91} \\
			Gaussian Noise ($\delta_n$=0.05) & 89.84 	& 97.06 								& 98.45 								& \textbf{99.64} \\ \hline
			Resize ($r_r$=50\%)				 & 98.70 	& \textbf{100} 							& 99.70 								& \textbf{100} \\
			Resize ($r_r$=30\%)				 & 97.86 	& \textbf{100} 							& 99.68 								& 99.99 \\
			Resize ($r_r$=10\%)				 & 88.70 	& 83.85 								& 97.58 								& \textbf{97.73} \\ \hline
			JPEG Compression ($q$=50)		 & 97.87 	& 98.97 								& 97.52 								& \textbf{99.72} \\
			JPEG Compression ($q$=30)		 & 91.91 	& 94.59 								& 92.46 								& \textbf{95.24} \\
			JPEG Compression ($q$=20)		 & 80.19 	& 81.23 								& \textbf{81.40} 						& 81.32 \\ 
			\hline \hline
		\end{tabular}
		\label{Table:RobustComparisonCN}
	\end{table*}

\subsection{Further improvement by adjusting embedding strength}\label{sec:improvement}

 Although the PSNR and SSIM values in Table~\ref{Table:QualityComparison} indicate a satisfying quality with the 100 bits of watermark, the examples in Fig.~\ref{fig:QualityComparison} show clear embedding traces at the background of document images. It can be attributed to the clean background of document image. We tried to alleviate this phenomenon by adjusting the embedding strength in training and testing stages.

 The adjustment of the embedding strength $\alpha$ will influence the quality of watermarked image and the bit accuracy according to the Formula~\ref{equ:addition}. As shown in Table~\ref{Table:RobustComparisonEN} and Table~\ref{Table:RobustComparisonCN}, our scheme achieves good bit accuracy even after the attacks with high strength. It indicates a space to decrease the embedding strength $\alpha$, so as to improve the quality of the watermarked image without much loss in bit accuracy. That is to say, we can set a relatively high embedding strength during the training process. After the model is fully trained, we can decrease the embedding strength in real application.
 
\begin{figure}[htbp]
	\includegraphics[width=\linewidth]{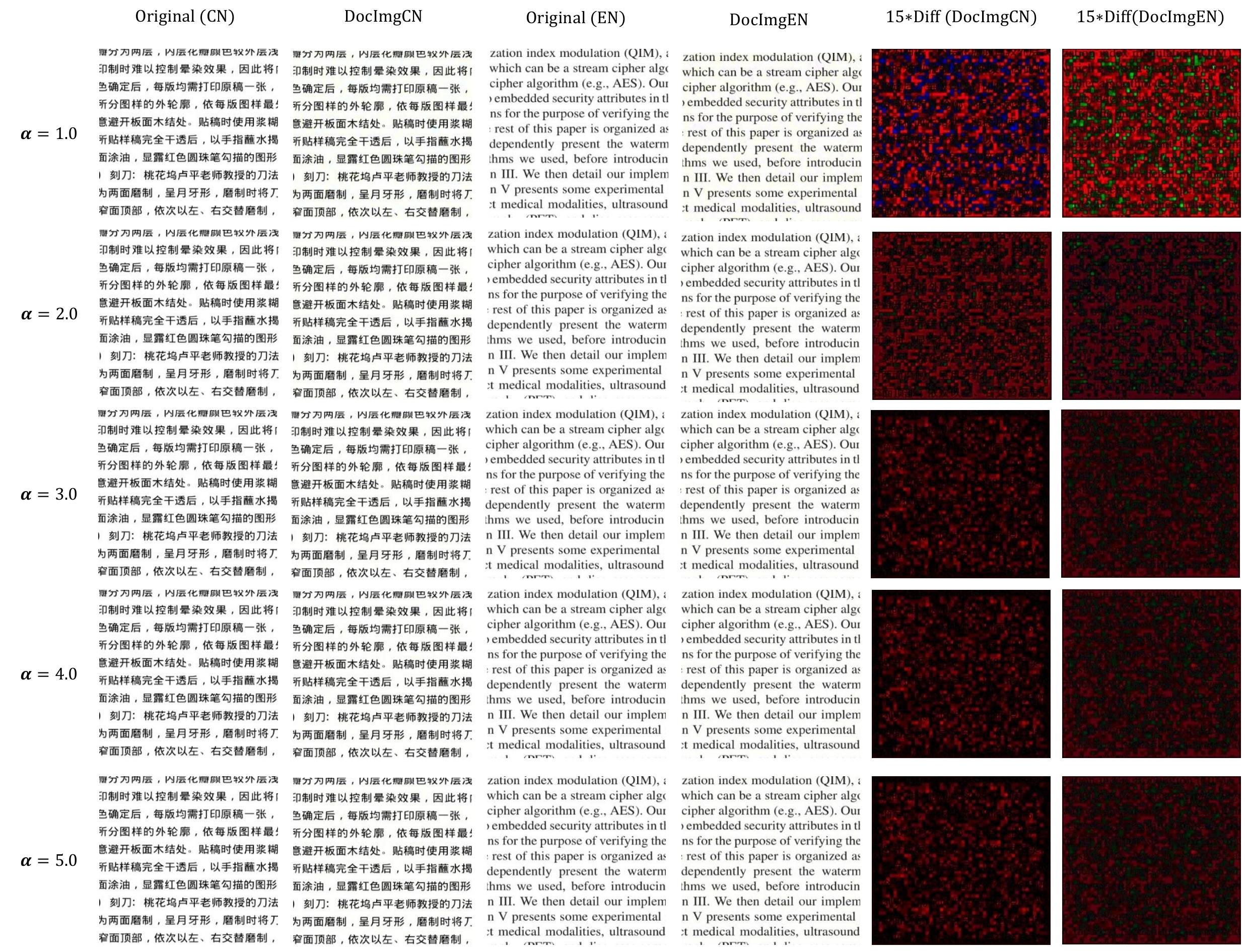}
	\caption{
		The visual examples of our scheme with different embedding strength during the testing stage.}
	\label{fig:QualityWithDifEmbeddingFactor}
 \end{figure}

 In this subsection, we try to set the embedding strength $\alpha=1.0, 2.0, 3.0, 4.0, 5.0$ during the training process and decrease $\alpha$ to be 1.0 in the testing stage. As shown in Table~\ref{Table:QualityWithDifEmbeddingFactor} and Fig.~\ref{fig:QualityWithDifEmbeddingFactor}, with larger initial $\alpha$, better image quality is achieved, while as shown in Table~\ref{Table:RobustWithDifEmbeddingFactorEN} and \ref{Table:RobustWithDifEmbeddingFactorCN}, the bit accuracy is not decreased a lot except for the JPEG-compressed images.
 
 \begin{table}[ht]
	\caption{
		The PSNR, SSIM and CPP of our scheme with different embedding strength during the testing stage.}
	\centering
	\begin{tabular}{c|ccc}
		\hline \hline
		Schemes							     & PSNR (dB)      & SSIM           & CPP \\ \hline
		On DocImgEN with $\alpha$=1.0        & 40.10          & 0.962          & 3.52 \\
		On DocImgEN with $\alpha$=2.0        & 44.80          & 0.990          & 2.07 \\
		On DocImgEN with $\alpha$=3.0        & 46.45          & 0.993          & \textbf{2.05} \\
		On DocImgEN with $\alpha$=4.0        & 46.80          & 0.995          & 2.58 \\
		On DocImgEN with $\alpha$=5.0        & \textbf{48.30} & \textbf{0.996} & 2.46 \\ \hline
		On DocImgCN with $\alpha$=1.0        & 41.07          & 0.965          & 3.77 \\
		On DocImgCN with $\alpha$=2.0        & 44.97          & 0.986          & 2.95 \\
		On DocImgCN with $\alpha$=3.0        & 45.23          & 0.992          & 2.91 \\
		On DocImgCN with $\alpha$=4.0        & 46.90          & 0.995          & \textbf{2.01} \\
		On DocImgCN with $\alpha$=5.0        & \textbf{47.30} & \textbf{0.996} & 2.60 \\ 
		\hline \hline
	\end{tabular}
	\label{Table:QualityWithDifEmbeddingFactor}
 \end{table}

 \begin{table*}[!htbp]
	\caption{
		Bit accuracy of our scheme with the different embedding strength during the testing stage on DocImgEN. }
	\centering
	\begin{tabular}{l|ccccc}
		\hline \hline
		\makecell[c]{Attack Type}	& DocImgEN ($\alpha$=1.0)	& DocImgEN ($\alpha$=2.0)	& DocImgEN ($\alpha$=3.0)	& DocImgEN ($\alpha$=4.0)	& DocImgEN ($\alpha$=5.0)\\ \hline
		Dropout ($r_d$=10\%)			 & 100				& 99.92			& 99.93						& 99.69						& 99.86\\
		Dropout ($r_d$=30\%)			 & 100				& 99.74			& 99.68						& 99.10						& 99.28\\
		Dropout ($r_d$=50\%)			 & 99.99			& 98.42			& 98.37						& 97.03						& 96.79\\ \hline
		Cropout ($r_c$=10\%)			 & 100				& 99.98			& 99.94						& 99.91						& 99.91\\
		Cropout ($r_c$=30\%)			 & 99.95			& 99.30			& 99.15						& 98.89						& 98.80\\
		Cropout ($r_c$=50\%)			 & 98.48			& 96.03			& 95.53						& 94.50						& 92.86\\ \hline
		Gaussian Blur ($\delta_b$=3)	 & 100				& 100			& 100						& 99.96						& 99.96\\
		Gaussian Blur ($\delta_b$=5)	 & 100				& 100			& 100						& 99.96						& 99.96\\
		Gaussian Blur ($\delta_b$=7)	 & 100				& 100			& 100						& 99.95						& 99.94\\ \hline
		Gaussian Noise ($\delta_n$=0.02) & 100				& 99.36 		& 99.21						& 98.29						& 97.57\\
		Gaussian Noise ($\delta_n$=0.03) & 100				& 97.76 		& 97.28						& 94.51						& 92.81\\
		Gaussian Noise ($\delta_n$=0.05) & 99.83			& 90.92 		& 90.44						& 83.99						& 78.80\\ \hline
		Resize ($r_r$=50\%)				 & 100				& 100			& 100						& 99.99						& 99.99\\
		Resize ($r_r$=30\%)				 & 99.99			& 100			& 99.99						& 99.97						& 99.92\\
		Resize ($r_r$=10\%)				 & 97.53			& 97.34			& 97.10						& 89.63						& 91.95\\ \hline
		JPEG Compression ($q$=50)		 & 99.81			& 89.17			& 88.28						& 76.99						& 73.94\\
		JPEG Compression ($q$=30)		 & 96.62			& 69.77			& 67.86						& 63.66						& 42.06\\
		JPEG Compression ($q$=20)		 & 88.68			& 42.43			& 38.89						& 38.02						& 30.44\\ 
		\hline \hline
	\end{tabular}
	\label{Table:RobustWithDifEmbeddingFactorEN}
 \end{table*}

 \begin{table*}[!htbp]
	\caption{
		Bit accuracy of our scheme with the different embedding strength during the testing stage on DocImgCN.}
	\centering
	\begin{tabular}{l|ccccc}
		\hline \hline
		\makecell[c]{Attack Type}	& DocImgCN ($\alpha$=1.0)	& DocImgCN ($\alpha$=2.0)	& DocImgCN ($\alpha$=3.0)	& DocImgCN ($\alpha$=4.0)	& DocImgCN ($\alpha$=5.0)\\ \hline
		Dropout ($r_d$=10\%)			 & 100				& 100			& 99.61 					& 99.31						& 99.24\\
		Dropout ($r_d$=30\%)			 & 99.96			& 99.94			& 99.21 					& 98.52						& 98.09\\
		Dropout ($r_d$=50\%)			 & 99.90			& 99.53			& 97.33 					& 95.41						& 94.04\\ \hline
		Cropout ($r_c$=10\%)			 & 99.99			& 100			& 99.68 					& 99.58						& 99.58\\
		Cropout ($r_c$=30\%)			 & 99.68			& 99.63			& 98.45 					& 98.52						& 97.97\\
		Cropout ($r_c$=50\%)			 & 97.99			& 96.28			& 92.44 					& 93.07 					& 92.78\\ \hline
		Gaussian Blur ($\delta_b$=3)	 & 100				& 100			& 100 						& 100						& 99.97\\
		Gaussian Blur ($\delta_b$=5)	 & 100				& 100			& 99.99 					& 100 						& 99.97\\
		Gaussian Blur ($\delta_b$=7)	 & 100				& 100			& 99.99 					& 100 						& 99.97\\ \hline
		Gaussian Noise ($\delta_n$=0.02) & 99.95			& 99.89			& 98.98 					& 95.69						& 95.13\\
		Gaussian Noise ($\delta_n$=0.03) & 99.91			& 99.38			& 97.08 					& 88.35						& 89.47\\
		Gaussian Noise ($\delta_n$=0.05) & 99.64			& 94.98			& 87.83 					& 71.16						& 76.56\\ \hline
		Resize ($r_r$=50\%)				 & 100				& 100			& 99.98 					& 99.97						& 99.97\\
		Resize ($r_r$=30\%)				 & 99.99			& 99.98			& 99.97 					& 99.91						& 99.95\\
		Resize ($r_r$=10\%)				 & 97.73			& 87.35			& 90.32 					& 86.60						& 91.41\\ \hline
		JPEG Compression ($q$=50)		 & 99.72			& 85.23			& 79.79 					& 69.46						& 64.84\\
		JPEG Compression ($q$=30)		 & 95.24			& 68.09			& 61.62 					& 57.83						& 46.16\\
		JPEG Compression ($q$=20)		 & 81.32			& 45.74			& 41.13 					& 41.09						& 40.42\\ 
		\hline \hline
	\end{tabular}
	\label{Table:RobustWithDifEmbeddingFactorCN}
 \end{table*}

\section{Conclusions}\label{sec:conclusion}

 This paper proposed an end-to-end document image watermarking scheme using deep neural network. An encoder and a decoder are designed to embed and extract the watermark. A noise layer is incorporated to simulate the various attacks such as the Cropout, Dropout, Gaussian Blur, Gaussian Noise, Resize, and JPEG Compression. Watermark is expanded to increase the robustness. The cover image and expanded watermark are repeatedly concatenated with the tensor during the training process to improve the performance. A text-sensitive loss function is designed to decrease the embedding modification on characters. An embedding strength adjustment strategy is further proposed to decrease the embedding trace with little loss of robustness. The extensive experiments demonstrated the superiority of our scheme. In future work, it could be promising to incorporate the embedding strength adjustment strategy in watermarking scheme for nature images. In addition, further efforts are needed to resist JPEG compression.

\section*{Acknowledgment}\label{sec:acknowledgment}

 This work is supported in part by the Jiangsu Basic Research Programs-Natural Science Foundation under grant numbers BK20181407, in part by the National Natural Science Foundation of China under grant numbers 62122032, 62102189, U1936118, and 61672294, in part by Six Peak Talent project of Jiangsu Province (R2016L13), Qinglan Project of Jiangsu Province, and “333” project of Jiangsu Province, in part by the Priority Academic Program Development of Jiangsu Higher Education Institutions (PAPD) fund, in part by the Collaborative Innovation Center of Atmospheric Environment and Equipment Technology (CICAEET) fund, China. Zhihua Xia is supported by BK21+ program from the Ministry of Education of Korea.

  \begin{IEEEbiography}[{\includegraphics[width=1in,height=1.25in,clip,keepaspectratio]{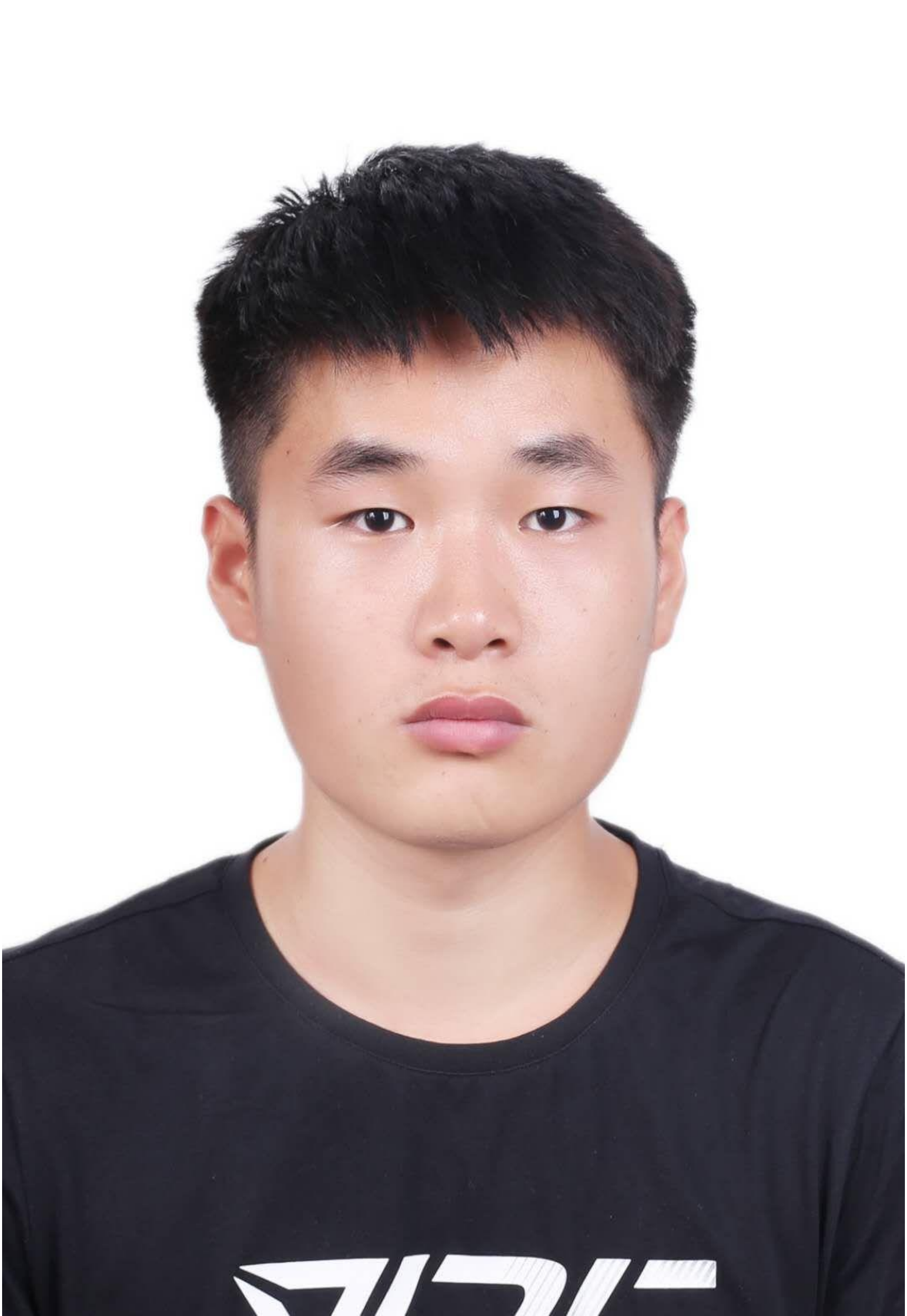}}]{Sulong Ge} received his BE degree in software engineering from TianGong University in 2020. He is currently pursuing master degree in School of Computer Science in Nanjing University of Information Science and Technology. His research interests include data hiding and information forensics.\end{IEEEbiography}

  \begin{IEEEbiography}[{\includegraphics[width=1in,height=1.25in,clip,keepaspectratio]{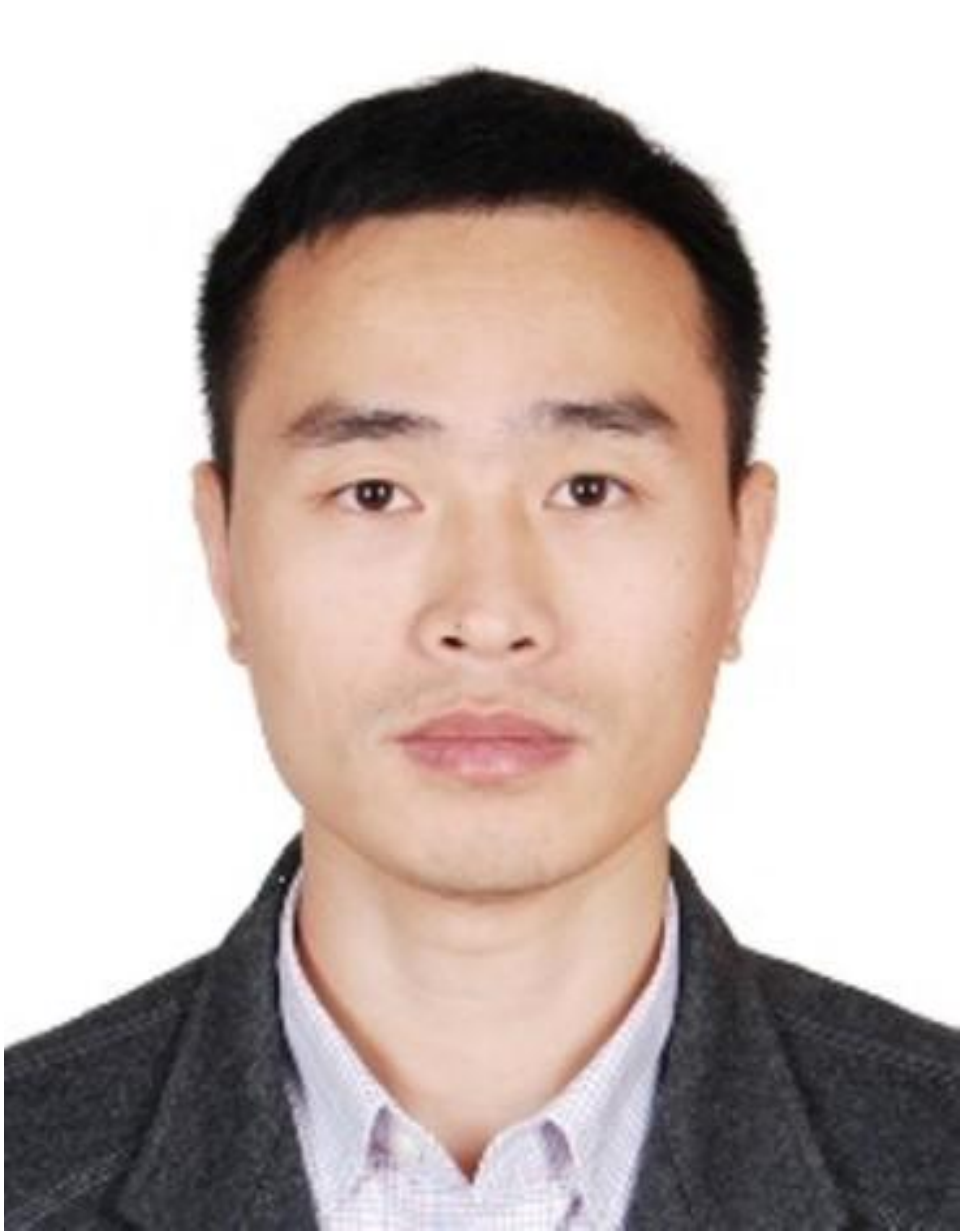}}]{Zhihua Xia} received his Ph.D. degree in computer science and technology from Hunan University, China, in 2011, and worked successively as a lecturer, an associate professor, and a professor with College of Computer and Software, Nanjing University of Information Science and Technology. He is currently a professor with the College of Cyber Security, Jinan University, China. He was a visiting scholar at New Jersey Institute of Technology, USA, in 2015, and was a visiting professor at Sungkyunkwan University, Korea, in 2016. He serves as a managing editor for IJAACS. His research interests include AI security, cloud computing security, and digital forensic. He is a member of the IEEE since Mar. 1, 2014.\end{IEEEbiography}

  \begin{IEEEbiography}[{\includegraphics[width=1in,height=1.25in,clip,keepaspectratio]{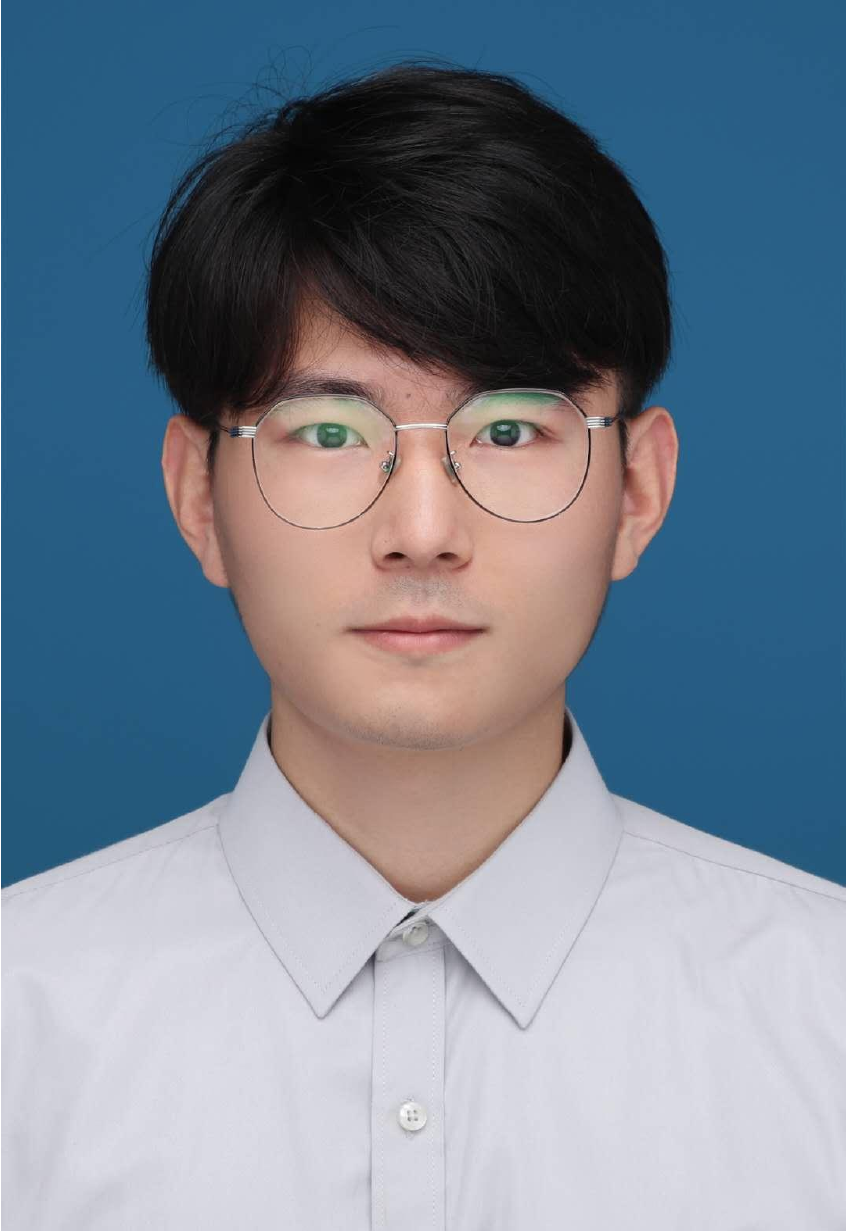}}]{Jianwei Fei} received his BE degree in Electronic and Information Engineering from Nanjing Forestry University in 2014. He is currently pursuing a master’s degree in Computer Science in Nanjing University of Information Science and Technology. His research interests include artificial intelligence security and multimedia forensics.\end{IEEEbiography}

  \begin{IEEEbiography}[{\includegraphics[width=1in,height=1.25in,clip,keepaspectratio]{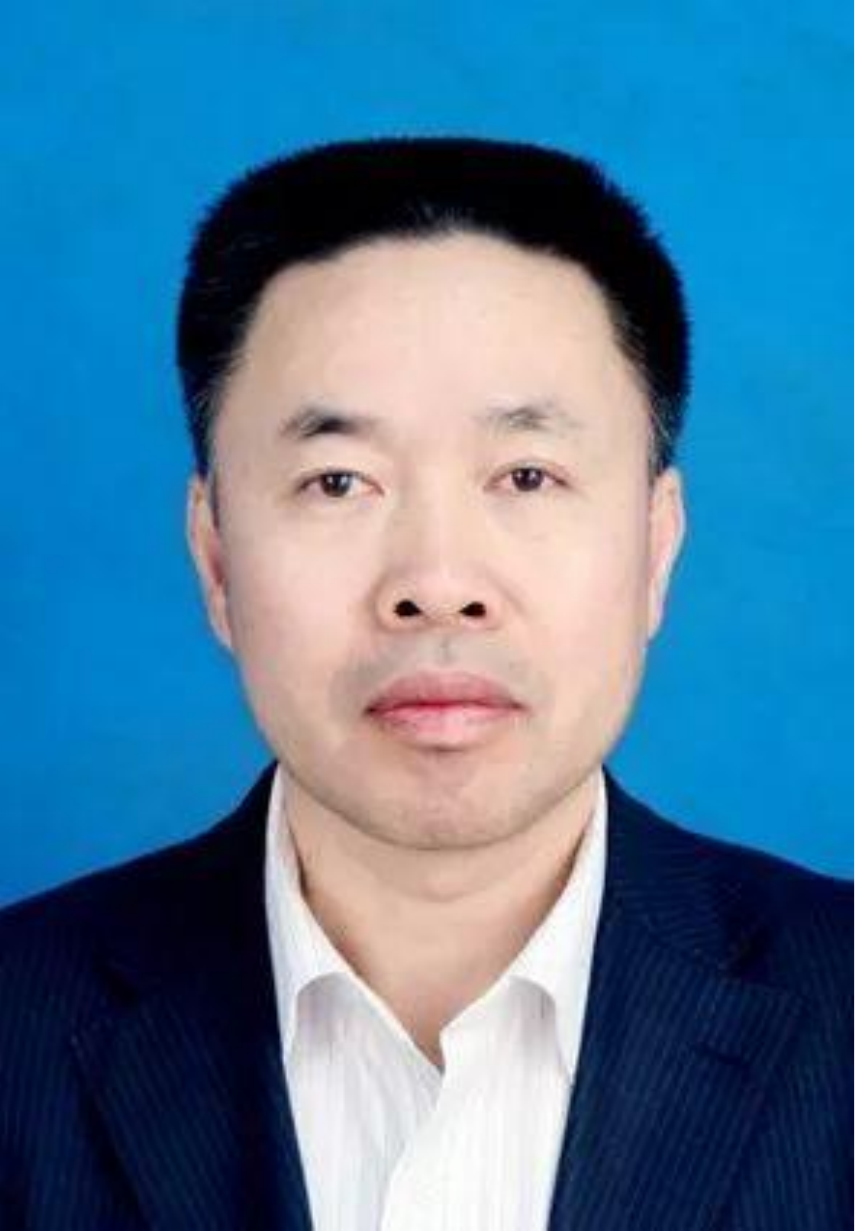}}]{Xingming Sun} (SM’07) received the B.S. degree in Mathematics from Hunan Normal University, China, in 1984, the M.S. degree in Computing Science from the Dalian University of Science and Technology, China, in 1998, and the Ph.D. degree in Computing Science from Fudan University, China, in 2001. He is currently a professor and the Dean in the College Computer and Software, Nanjing University of Information Science and Technology, China. In 2006, he visited University College London, UK. He was a visiting professor with the University of Warwick, UK, in 2008 and 2010. His research interests include network and information security, digital watermarking, cloud computing security, and wireless network security. He is the general chair of ICCCS (International Conference of Cloud Computing and Security) 2015, 2016, 2017, and 2018. His research has been supported by NSFC, 863, 973. He is a recipient of Science and Technology Progress Award, and a senior member of IEEE.\end{IEEEbiography}

  \begin{IEEEbiography}[{\includegraphics[width=1in,height=1.25in,clip,keepaspectratio]{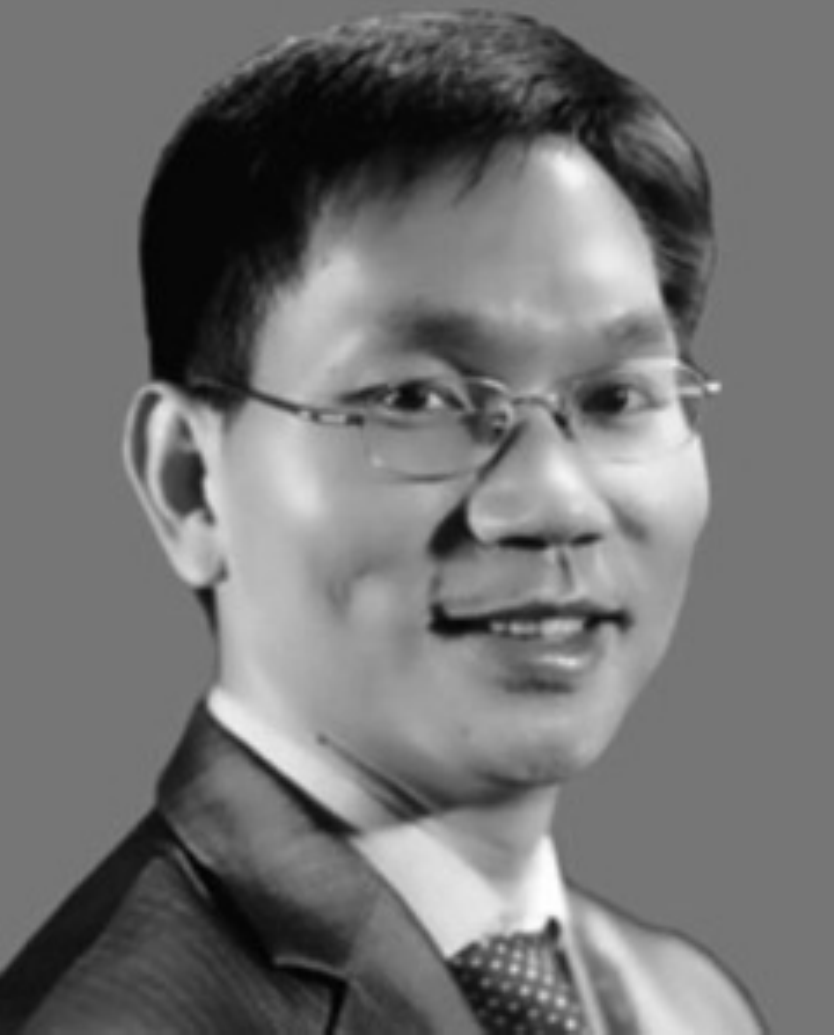}}]{Jian Weng} received the Ph.D. degree in computer science and engineering from Shanghai Jiao Tong University, Shanghai, China, in 2008. He is currently a Professor and the Dean with the College of Information Science and Technology, Jinan University, Guangzhou, China. His research interests include public key cryptography, cloud security, and blockchain. He was the PC Co-Chairs or PC Member for more than 30 international conferences. He also serves as an Associate Editor for the IEEE TRANSACTIONS ON VEHICULART ECHNOLOGY.\end{IEEEbiography}


\begin{thebibliography}{10}

  \bibitem{cox2007digital}
  I.~Cox, M.~Miller, J.~Bloom, J.~Fridrich, and T.~Kalker, \emph{Digital
    watermarking and steganography}.\hskip 1em plus 0.5em minus 0.4em\relax
    Morgan kaufmann, 2007.
  
  \bibitem{fang2019camera}
  H.~Fang, W.~Zhang, Z.~Ma, H.~Zhou, S.~Sun, H.~Cui, and N.~Yu, ``A camera
    shooting resilient watermarking scheme for underpainting documents,''
    \emph{IEEE Transactions on Circuits and Systems for Video Technology},
    vol.~30, no.~11, pp. 4075--4089, 2019.
  
  \bibitem{brassil1994electronic}
  J.~Brassil, S.~Low, N.~Maxemchuk, and L.~O'Gorman, ``Electronic marking and
    identification techniques to discourage document copying,'' in
    \emph{Proceedings of INFOCOM'94 Conference on Computer Communications}.\hskip
    1em plus 0.5em minus 0.4em\relax IEEE, 1994, pp. 1278--1287.
  
  \bibitem{brassil1999copyright}
  J.~Brassil, S.~Low, and N.~Maxemchuk, ``Copyright protection for the electronic
    distribution of text documents,'' \emph{Proceedings of the IEEE}, vol.~87,
    no.~7, pp. 1181--1196, 1999.
  
  \bibitem{amano1999feature}
  T.~Amano and D.~Misaki, ``A feature calibration method for watermarking of
    document images,'' in \emph{Proceedings of the Fifth International Conference
    on Document Analysis and Recognition. ICDAR'99 (Cat. No. PR00318)}.\hskip 1em
    plus 0.5em minus 0.4em\relax IEEE, 1999, pp. 91--94.
  
  \bibitem{huang2001interword}
  D.~Huang and H.~Yan, ``Interword distance changes represented by sine waves for
    watermarking text images,'' \emph{IEEE Transactions on Circuits and Systems
    for Video Technology}, vol.~11, no.~12, pp. 1237--1245, 2001.
  
  \bibitem{kim2003text}
  Y.-W. Kim, K.-A. Moon, and I.-S. Oh, ``A text watermarking algorithm based on
    word classification and inter-word space statistics.'' in \emph{ICDAR}.\hskip
    1em plus 0.5em minus 0.4em\relax Citeseer, 2003, pp. 775--779.
  
  \bibitem{tan2012printscan}
  L.~Tan, X.~Sun, and G.~Sun, ``Print-scan resilient text image watermarking
    based on stroke direction modulation for chinese document authentication.''
    \emph{Radioengineering}, vol.~21, no.~1, 2012.
  
  \bibitem{kamaruddin2018review}
  N.~S. Kamaruddin, A.~Kamsin, L.~Y. Por, and H.~Rahman, ``A review of text
    watermarking: theory, methods, and applications,'' \emph{IEEE Access},
    vol.~6, pp. 8011--8028, 2018.
  
  \bibitem{kim2004watermarking}
  Y.-W. Kim and I.-S. Oh, ``Watermarking text document images using edge
    direction histograms,'' \emph{Pattern Recognition Letters}, vol.~25, no.~11,
    pp. 1243--1251, 2004.
  
  \bibitem{loc2018document}
  C.~V. Loc, J.-C. Burie, and J.-M. Ogier, ``Document images watermarking for
    security issue using fully convolutional networks,'' in \emph{2018 24th
    International conference on pattern recognition (ICPR)}.\hskip 1em plus 0.5em
    minus 0.4em\relax IEEE, 2018, pp. 1091--1096.
  
  \bibitem{lu2002watermark}
  H.~Lu, X.~Shi, Y.~Q. Shi, A.~C. Kot, and L.~Chen, ``Watermark embedding in dc
    components of dct for binary images,'' in \emph{2002 IEEE Workshop on
    Multimedia Signal Processing.}\hskip 1em plus 0.5em minus 0.4em\relax IEEE,
    2002, pp. 300--303.
  
  \bibitem{rosiyadi2011copyright}
  D.~Rosiyadi, S.-J. Horng, P.~Fan, X.~Wang, M.~K. Khan, and Y.~Pan, ``Copyright
    protection for e-government document images,'' \emph{IEEE MultiMedia},
    vol.~19, no.~3, pp. 62--73, 2011.
  
  \bibitem{chetanK2015efficient}
  K.~Chetan and S.~Nirmala, ``An efficient and secure robust watermarking scheme
    for document images using integer wavelets and block coding of binary
    watermarks,'' \emph{Journal of Information Security and Applications},
    vol.~24, pp. 13--24, 2015.
  
  \bibitem{al2017copyright}
  A.~Al-Haj and H.~Barouqa, ``Copyright protection of e-government document
    images using digital watermarking,'' in \emph{2017 3rd International
    Conference on Information Management (ICIM)}.\hskip 1em plus 0.5em minus
    0.4em\relax IEEE, 2017, pp. 441--446.
  
  \bibitem{dang2019blind}
  Q.~B. Dang, K.~Louisa, M.~Coustaty, M.~M. Luqman, and J.-M. Ogier, ``A blind
    document image watermarking approach based on discrete wavelet transform and
    qr code embedding,'' in \emph{2019 International Conference on Document
    Analysis and Recognition Workshops (ICDARW)}, vol.~8.\hskip 1em plus 0.5em
    minus 0.4em\relax IEEE, 2019, pp. 1--6.
  
  \bibitem{mun2019finding}
  S.-M. Mun, S.-H. Nam, H.~Jang, D.~Kim, and H.-K. Lee, ``Finding robust domain
    from attacks: A learning framework for blind watermarking,''
	\emph{Neurocomputing}, vol. 337, pp. 191--202, 2019.
  
  \bibitem{ahmadi2020redmark}
  M.~Ahmadi, A.~Norouzi, N.~Karimi, S.~Samavi, and A.~Emami, ``Redmark: Framework
    for residual diffusion watermarking based on deep networks,'' \emph{Expert
	Systems with Applications}, vol. 146, p. 113157, 2020.
  
  \bibitem{zhong2020automated}
  X.~Zhong, P.-C. Huang, S.~Mastorakis, and F.~Shih, ``An automated and robust
    image watermarking scheme based on deep neural networks,'' \emph{IEEE
	Transactions on Multimedia}, vol.~23, pp. 1951--1961, 2021.
  
  \bibitem{zhu2018hidden}
  J.~Zhu, R.~Kaplan, J.~Johnson, and L.~Fei-Fei, ``Hidden: Hiding data with deep
    networks,'' in \emph{Proceedings of the European conference on computer
    vision (ECCV)}, 2018, pp. 657--672.
  
  \bibitem{liu2019novel}
  Y.~Liu, M.~Guo, J.~Zhang, Y.~Zhu, and X.~Xie, ``A novel two-stage separable
    deep learning framework for practical blind watermarking,'' in
    \emph{Proceedings of the 27th ACM International Conference on Multimedia},
    2019, pp. 1509--1517.
  
  \bibitem{luo2020distortion}
  X.~Luo, R.~Zhan, H.~Chang, F.~Yang, and P.~Milanfar, ``Distortion agnostic deep
    watermarking,'' in \emph{Proceedings of the IEEE/CVF Conference on Computer
    Vision and Pattern Recognition}, 2020, pp. 13\,548--13\,557.
  
  \bibitem{shin2017jpeg}
  R.~Shin and D.~Song, ``Jpeg-resistant adversarial images,'' in \emph{NIPS 2017
    Workshop on Machine Learning and Computer Security}, vol.~1, 2017.
  
  \bibitem{luo2021rate}
  X.~Luo, H.~Talebi, F.~Yang, M.~Elad, and P.~Milanfar, ``The
    rate-distortion-accuracy tradeoff: Jpeg case study,'' \emph{2021 Data
    Compression Conference (DCC)}, pp. 354--354, 2021.
  
  \bibitem{krizhevsky2012imagenet}
  A.~Krizhevsky, I.~Sutskever, and G.~E. Hinton, ``Imagenet classification with
    deep convolutional neural networks,'' \emph{Advances in neural information
	processing systems}, vol.~25, pp. 1097--1105, 2012.
  
  \bibitem{zhao2016loss}
  H.~Zhao, O.~Gallo, I.~Frosio, and J.~Kautz, ``Loss functions for image
    restoration with neural networks,'' \emph{IEEE Transactions on computational
    imaging}, vol.~3, no.~1, pp. 47--57, 2016.
  
  \bibitem{ieee2021database}
  ``Ieee/iet electronic library,'' [Online], accessed: April 4, 2021. Available:
	\url{https://ieeexplore.ieee.org/Xplore/home.jsp}.
  
  \bibitem{cnki2021database}
  ``China national knowledge infrastructure,'' [Online], accessed: April 4, 2021.
    Available: \url{https://www.cnki.net/}.
  
  \bibitem{kingma2014adam}
  D.~P. Kingma and J.~Ba, ``Adam: A method for stochastic optimization,''
	\emph{arXiv preprint arXiv:1412.6980}, 2014.

  \bibitem{wang2004image}
  Z.~Wang, A.~C. Bovik, H.~R. Sheikh, and E.~P. Simoncelli, ``Image quality
    assessment: from error visibility to structural similarity,'' \emph{IEEE
    transactions on image processing}, vol.~13, no.~4, pp. 600--612, 2004.
  
  \bibitem{tancik2020stegastamp}
  M.~Tancik, B.~Mildenhall, and R.~Ng, ``Stegastamp: Invisible hyperlinks in
    physical photographs,'' in \emph{Proceedings of the IEEE/CVF Conference on
    Computer Vision and Pattern Recognition}, 2020, pp. 2117--2126.
  
  \bibitem{zhang2018unreasonable}
  R.~Zhang, P.~Isola, A.~A. Efros, E.~Shechtman, and O.~Wang, ``The unreasonable
    effectiveness of deep features as a perceptual metric,'' in \emph{Proceedings
    of the IEEE conference on computer vision and pattern recognition}, 2018, pp.
    586--595.
  \end{thebibliography}
\end{document}